\documentclass{bmvc2k}

\usepackage{amsmath,amssymb}
\usepackage{multirow}
\usepackage{wrapfig}
\usepackage{color}
\usepackage{colortbl}
\usepackage{tabularx}
\usepackage{url}
\usepackage{float}
\usepackage{booktabs}
\usepackage{hyperref}
\usepackage{caption}
\usepackage{bbding}
\usepackage{xcolor}
\usepackage{array}

\title{Widely Applicable Strong Baseline for Sports Ball Detection and Tracking}

\addauthor{Shuhei Tarashima}{tarashima@acm.org}{1,2}
\addauthor{Muhammad Abdul Haq}{muhabdulhaq@gmail.com}{2}
\addauthor{Yushan Wang}{yushanwang218@gmail.com}{2}
\addauthor{Norio Tagawa}{tagawa@tmu.ac.jp}{2}

\addinstitution{
 Innovation Center\\
 NTT Communications Corporation\\
 Tokyo, Japan
}
\addinstitution{
 Faculty of Systems Design\\
 Tokyo Metropolitan University\\
 Tokyo, Japan
}

\runninghead{Tarashima \etal}{Widely Applicable Strong Baseline for SBDT}

\def\eg{\emph{e.g}\bmvaOneDot}

\def\etal{\emph{et al}\bmvaOneDot}
\def\ie{\emph{i.e}\bmvaOneDot}

\begin{document}

\maketitle

\begin{abstract}
In this work, we present a novel Sports Ball Detection and Tracking (SBDT) method that can be applied to various sports categories.
Our approach is composed of (1) high-resolution feature extraction, (2) position-aware model training, and (3) inference considering temporal consistency, all of which are put together as a new SBDT baseline.
Besides, to validate the wide-applicability of our approach, we compare our baseline with 6 state-of-the-art SBDT methods on 5 datasets from different sports categories.
We achieve this by newly introducing two SBDT datasets, providing new ball annotations for two datasets, and re-implementing all the methods to ease extensive comparison.
Experimental results demonstrate that our approach is substantially superior to existing methods on all the sports categories covered by the datasets.
We believe our proposed method can play as a Widely Applicable Strong Baseline (WASB) of SBDT, and our datasets and codebase will promote future SBDT research.
Datasets and codes are available at \url{https://github.com/nttcom/WASB-SBDT}.
\end{abstract}
\section{Introduction}
\label{sec:intro}
\begin{figure*}[t]
\centering
\includegraphics[width=1.0\textwidth, page=2]{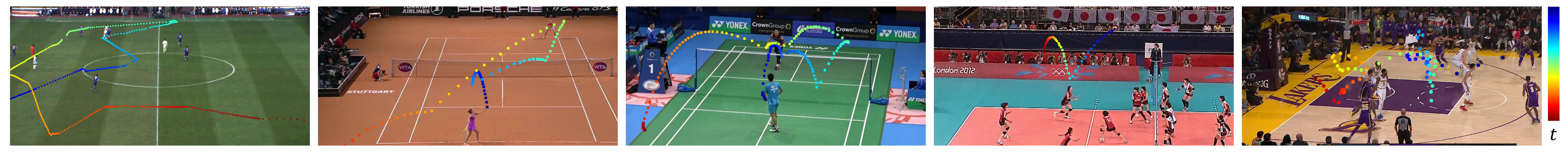}
\caption{Exemplar ball trajectories extracted from soccer, tennis, badminton, volleyball and basketball videos, respectively. Best viewed in color.}
\label{fig:task}
\vspace*{-5mm}
\end{figure*}

Sports ball trajectory depicted in Figure \ref{fig:task} is an important statistic for analytics of various sports such as
badminton \cite{wang+2022aaai},
baseball \cite{shum+2004icme},
basketball \cite{fu+2011vcip},
golf \cite{huang+2012icppw},
soccer \cite{theagarajan+cvpr2018,sarkar+2019cvprw},
tennis \cite{pingali+2000icpr},
table tennis \cite{desai+2005prmi},
and
volleyball \cite{cheng+2016icassp}.
Several commercial systems like Hawk-Eye\footnote{\url{https://www.hawkeyeinnovations.com/track}} and KINEXON\footnote{\url{https://kinexon.com/technology/ball-tracking/}} have already been successfully introduced to professional leagues, but they usually require high-cost installation.
Computer vision techniques can be an alternative approach to obtain ball trajectories from easily available video data.
However, this Sports Ball Detection and Tracking (SBDT) task is challenging due to the small size of a sports ball, its high speed, occlusion, blending in with surroundings, and camera motion \cite{yu+2003acmmm}.
\par
This SBDT task can uniformly be defined through various ball-games.
Therefore, {\it wide applicability} is an important property to be equipped by good SBDT methods.
However, while there are extensive literatures of SBDT methods proposed in the last two decades, most of them cannot be directly applied to different domains, since they are tailor-made for specific sports
(\eg,
badminton \cite{chen+2007tst},
baseball \cite{shum+2004icme},
basketball \cite{chen+2009jvcir,chakraborty+2011sic,chakraborty+2012icspcc,chakraborty+2013indicon,chakraborty+2013jo},
golf \cite{lyu+2015icia,lyu+2017ijsr},
soccer \cite{ohno+1999mfi,ohno+2000icpr,yu+2003acmmm,yu+2003icme,yu+2003icme2,choi+2004smvp,tong+2004icpr,yu+2004acmmm,choi+2005iciap,li+2005avss,liang+2005pcm,yu+2005icme,liu+2006ivc,ren+2006eccvw,shimawaki+2006icpr,yu+2006tmm,ishii+2007pcm,liang+2007tce,misu+2007icassp,yu+2007icme,yu+2007avc,ariki+2008icme,huang+2008icpr,pallavi+2008jvcir,ren+2008tcsvt,zhu+2008civr,beetz+2009ijcss,dorazio+2009avss,dorazio+2009tcsvt,dorazio+2009cviu,kim+2009cgiv,miura+2009cviu,ren+2009cviu,zhu+2009tmm,rao+2015iccsp},
tennis \cite{pingali+2000icpr,lepetit+2003cvpr,yu+2004icip,kittler+2005ia,yan+2005bmvc,kittler+2007iciap,yu+2007vcip,ekinci+2008vis,yan+2008tpami,conaire+2009icdsp,yu+2009cviu,teachabarikiti+2010icca,wong+2010ics,huang+2011avsp,almajai+2012dir,huang+2012apsipa,zhou+2013icassp,yan+2014cvs,archana+2015pcs,zhou+2015tmm,reno+2016tishw,yuan+2016scis},
table tennis \cite{zaveri+2004icme,desai+2005prmi,abed+2006acivs,chen+2006cesa,zhang+2010tim,zhang+2011zus,glover+2014icra,myint+2015mva},
volleyball \cite{chen+2007icassp,chen2012mta,cheng+2015pcm,cheng+2016icassp}).
Recent approaches \cite{komorowski+2019mva,komorowski+2020visapp,zandycke+2019mmsports,huang+2019avss,sun+2020icpai,liu+2022cvprw} based on Convolutional Neural Networks (CNNs) can potentially be used for different ball-games, but unfortunately in their works evaluations are limited to almost one sports category.
\par
Here we aim at building a new state-of-the-art (SOTA) SBDT method widely applicable to various sports categories.
To achieve this goal, we will make the following contributions:
\begin{itemize}
    \item While current SOTAs \cite{komorowski+2019mva,komorowski+2020visapp,zandycke+2019mmsports,huang+2019avss,sun+2020icpai,liu+2022cvprw} successfully solve the SBDT task on limited sports domains, we found that there is room for improvement with respect to (1) high-resolution feature extraction, (2) model training being aware of tiny ball position, and (3) inference which takes temporal consistency of ball position into account. We propose a series of solutions to ameliorate these drawbacks of existing methods, and put them together into a new SBDT approach.
    \item Different from the most SBDT works that evaluate their methods for almost one sports category, we use 5 datasets from different sports categories (\ie, badminton, basketball, soccer, tennis, volleyball) to compare our approach with 6 SOTA SBDT methods \cite{komorowski+2019mva,zandycke+2019mmsports,sun+2020icpai,liu+2022cvprw}. We establish this experimental protocol by introducing two novel datasets, providing new manual annotations for two datasets, and re-implementing all the existing methods. Experimental results demonstrate that our method substantially outperforms all the SBDT methods on all the datasets used in our evaluation.
\end{itemize}
\noindent
These contributions indicate that our proposed approach can play as a Widely Applicable Strong Baseline (WASB) of SBDT.
Also, we make datasets and codebases publicly available, which we believe promotes future SBDT research.
\section{Related Work}
\label{sec:related}
Roughly speaking, classical SBDT methods
\cite{lepetit+2003cvpr,yu+2003acmmm,yu+2003icme,yu+2003icme2,tong+2004icpr,yu+2004icip,zaveri+2004icme,desai+2005prmi,kittler+2005ia,liang+2005pcm,yan+2005bmvc,abed+2006acivs,chen+2006cesa,ren+2006eccvw,shimawaki+2006icpr,yu+2006tmm,chen+2007tst,chen+2007icassp,yu+2007icme,yu+2007avc,yu+2007vcip,ariki+2008icme,huang+2008icpr,pallavi+2008jvcir,yan+2008tpami,zhu+2008civr,beetz+2009ijcss,zhu+2009tmm,kim+2009cgiv,miura+2009cviu,wong+2010ics,teachabarikiti+2010icca,chakraborty+2011sic,chakraborty+2012icspcc,chakraborty+2013indicon,chakraborty+2013jo,zhou+2013icassp,yan+2014cvs,archana+2015pcs,lyu+2015icia,zhou+2015tmm,lyu+2017ijsr}
are based on {\it tracking-by-detection} paradigm:
Ball candidates are first detected from each video frame, then a true trajectory is recovered by associating the candidates through time.
The most typical ball candidate detector is a temporal background subtraction.
However, this approach can easily be contaminated by non-ball moving objects like players, even though it requires careful tuning to the target domain.
\par
Recent methods \cite{reno+2018cvprw,kamble+2018oer,voeikov+2020cvprw,komorowski+2020visapp,komorowski+2019visapp,komorowski+2019mva,zandycke+2019mmsports,wu+2020iet,ghasemzadeh+2021bmvc,huang+2019avss,sun+2020icpai,liu+2022cvprw} significantly ameliorate the above issue by employing encoder-decoder CNN models.
For example, DeepBall \cite{komorowski+2019visapp,komorowski+2019mva} is composed of a variant of fully convolutional networks \cite{liu+2016eccv}, in which intermediate multi-scale features are fused in a decoder to extract high-resolution heatmaps representing ball positions.
BallSeg \cite{zandycke+2019mmsports} is a modification of ICNet \cite{zhao+2018eccv}, so that two consecutive frames can be fed into the model to capture ball dynamics.
TrackNet and its variants \cite{huang+2019avss,sun+2020icpai,liu+2022cvprw} are based on U-Net \cite{ronneberger+2015miccai} architecture, following a multiple-in multiple-out (MIMO) design to efficiently capture ball movement.
Usually, training these models inevitably confronts high foreground-background class imbalance, due to the small ball size appeared in sports videos.
Existing methods address this issue by adapting the focal loss \footnote{The WBCE loss proposed in \cite{sun+2020icpai} is equivalent to the focal loss.} \cite{lin+2017iccv},
the combo loss \cite{taghanaki+cmig} or hard negative mining technique \cite{liu+2016eccv}.
Notice that in these recent methods, ball dynamics are considered only within frames that are combined in the same batch.
\par
We argue that, in recent methods described above, there is room for improvement with respect to (1) high-resolution feature extraction, (2) model training being aware of tiny ball position, and (3) inference which takes temporal consistency of ball position into account.
In the next section, we introduce solutions to improve these potential drawbacks.
\section{Widely Applicable Strong Baseline (WASB)}
\label{sec:method}
Following the majority of the SBDT literature\footnote{Some exceptional works like \cite{zhang+2022arxiv} define a ball position as a bounding box.}, our goal is to detect a $(x,y)$-coordinate of ball location from each image in a given video clip.
Similar to the recent works \cite{komorowski+2019mva,komorowski+2019visapp,zandycke+2019mmsports,sun+2020icpai,huang+2019avss,liu+2022cvprw}, we solve this problem by training a neural network that predicts heatmaps representing ball positions in input images.
At inference time, ball positions are determined by post-processing the heatmaps.
In the followings we detail our model, training and inference, all of which are put together into our proposed Widely Applicable Strong Baseline (WASB) for SBDT.
\subsection{High-Resolution Feature Extraction Model}
\label{sec:method:cnn}
Here we build a model that can produce heatmaps of the same spatial resolution $H \times W$ with an input tensor.
Recent works \cite{komorowski+2019mva,komorowski+2019visapp,zandycke+2019mmsports,sun+2020icpai,huang+2019avss,liu+2022cvprw} demonstrate the importance of a high-resolution and semantically-rich feature representation to precisely detect tiny sports balls.
In their methods, heatmaps are generated by combining highly-semantic but low-resolution decoder outputs with intermediate features produced by encoders to complement their spatial resolution.
We argue that, however, this encoder-decoder architecture can be a drawback for SBDT, since features to be combined lack one of the two required perspectives.
\par
\begin{wrapfigure}{r}[0pt]{0.4\textwidth}
\vspace*{-5mm}
\centering
\includegraphics[width=0.4\textwidth, page=2]{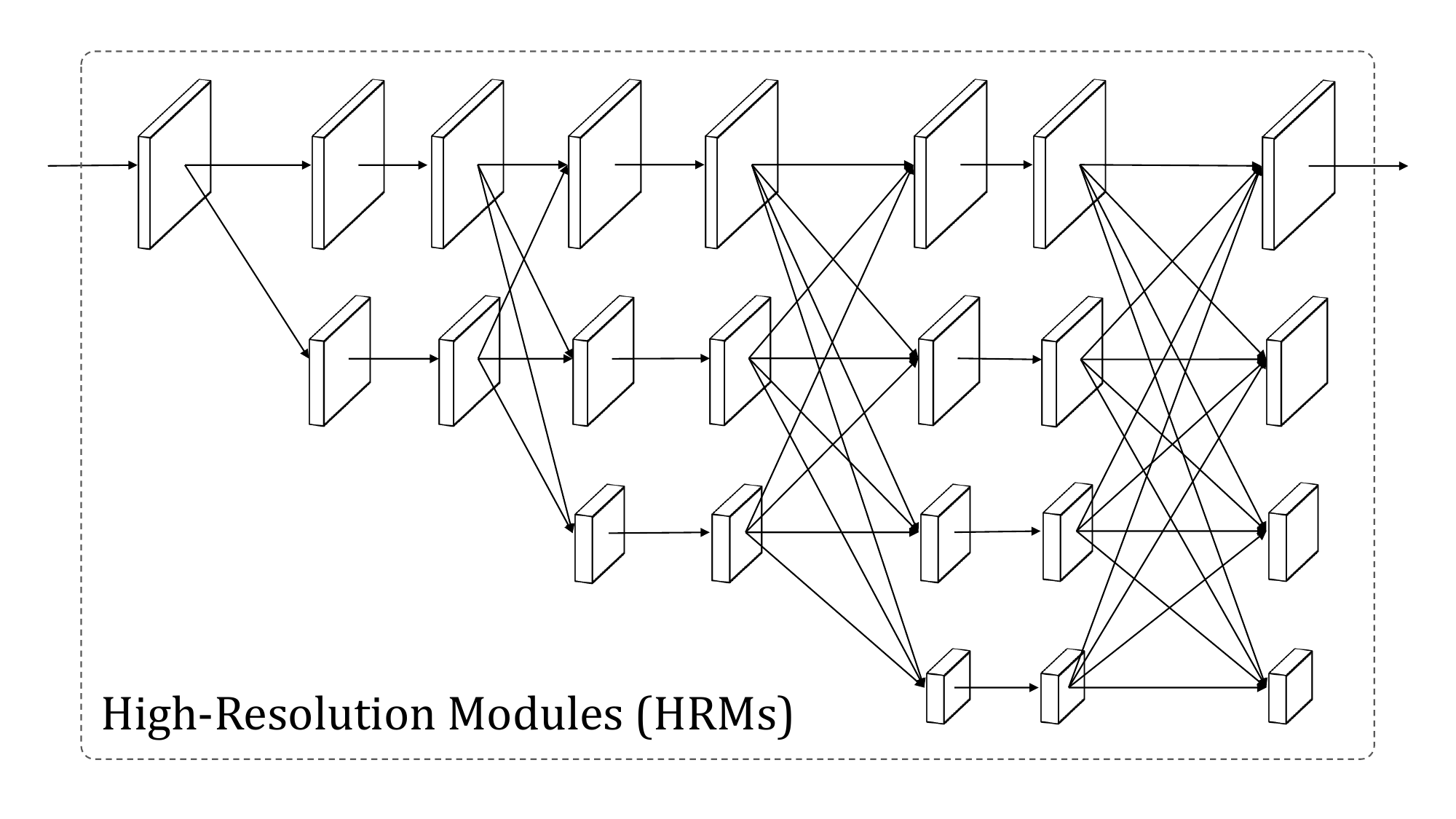}
\caption{High-Resolution Modules (HRMs) of our SBDT method.}
\label{fig:hrms}
\end{wrapfigure}
Based on this observation, in this work we propose to employ a CNN module that can produce semantically-rich representation without losing spatial resolution:
Specifically, we adopt a high-resolution feature extraction method proposed by a series of HRNet works \cite{wang+2021tpami,yu+2021cvpr}.
HRNet consists of a stem block and multi-stage high-resolution modules (HRMs), where in each new stage one high-to-low resolution convolution block is incrementally added.
The information across resolutions is exchanged repeatedly, which allows us to obtain a highly-semantic representation while keeping spatial resolution.
In this paper we instantiate our HRMs following the small HRNet design\footnote{\url{https://github.com/HRNet/HRNet-Image-Classification}} illustrated in Figure \ref{fig:hrms}:
There are 4 stages and each stage consists of parallel sequences of residual blocks \cite{he+2016cvpr} followed by a multi-resolution fusion.
\par
\begin{figure*}[t]
\centering
\includegraphics[width=1.0\textwidth, page=2]{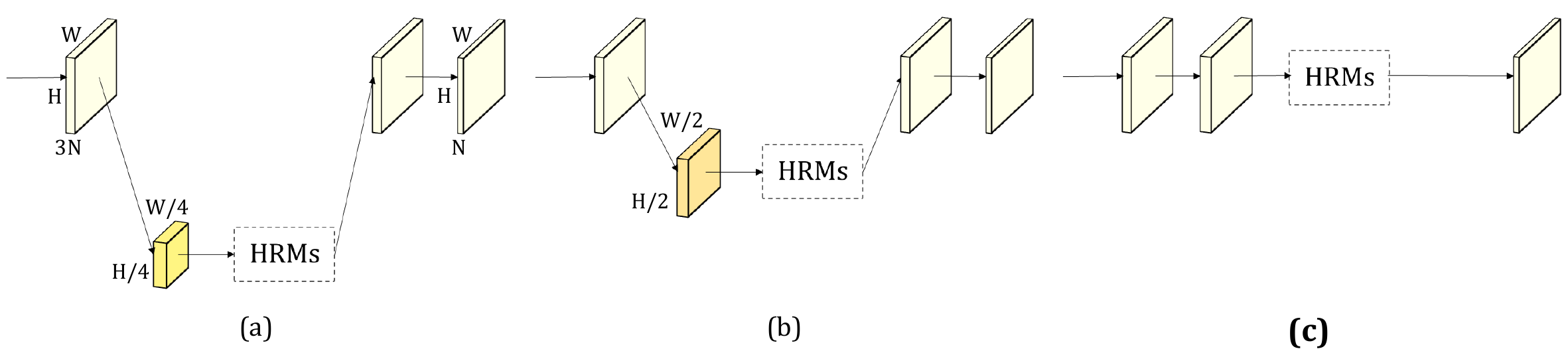}
\caption{(a) In the original stem design of HRNet \cite{wang+2021tpami,yu+2021cvpr}, spatial resolution of an input is reduced to one-fourth to be fed into HRMs. Alternatively, we propose to remove strides from the stem so that the resolution of intermediate features to be higher, as shown in (b) and (c). $N$ is the number of frames. We use (c) by default based on the ablation result in \S \ref{sec:eval:ablation}.}
\label{fig:arch}
\vspace*{-5mm}
\end{figure*}

If we directly follow the HRNet \cite{wang+2021tpami,yu+2021cvpr}, the feature fed into HRMs is down-sized to one-fourth by the stem block ({\it cf.} Figure \ref{fig:arch} (a)).
To make the resolution of intermediate representations higher, we propose to remove strides from the stem block and feed a tensor with higher spatial resolution to the HRMs, which are illustrated in Figure \ref{fig:arch} (b) and (c).
Notice that computational complexity increases when strides are removed.
We specifically adopt the model shown in Figure \ref{fig:arch} (c) by default, since it achieves higher SBDT performance with reasonable sacrifice of inference efficiency ({\it cf.} \S \ref{sec:eval:ablation}).
\par
To capture temporal dynamics of fast-moving sports balls, we follow the MIMO design like \cite{sun+2020icpai,liu+2022cvprw}:
$N$ consecutive frames are concatenated along the channel dimension then the resulting $H \times W \times 3N$ tensor is fed into our model, which generates the corresponding $N$ heatmaps of the same spatial resolution with the input (\ie, $H \times W \times N$).
\subsection{Position-Aware Model Training}
\label{sec:method:train}
To train SBDT models, we need to prepare ground truth (GT) maps from 2D ball positions, then optimize the model parameters by minimizing a loss between model predictions and GT maps.
Given a GT ball position
$\mathbf{p}^{GT} \in \mathbb{R}^{2}$
in an image, existing methods \cite{komorowski+2019mva,komorowski+2019visapp,zandycke+2019mmsports,sun+2020icpai,huang+2019avss,liu+2022cvprw} generate a {\it binary} GT map $\mathbf{y}^{bin}$ based on the following Equation \ref{eq:hm1}:
\begin{wrapfigure}{r}[0pt]{0.5\textwidth}
\vspace*{-2mm}
\centering
\begin{tabular}{cc}
\begin{minipage}[t]{0.2\textwidth}
\centering
\includegraphics[width=0.9\textwidth, page=2]{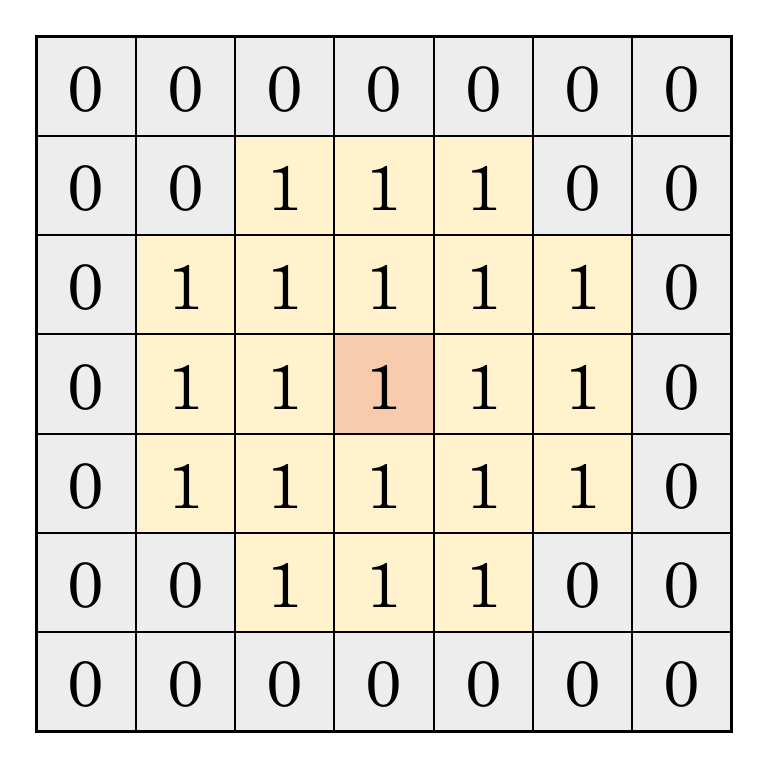}
\end{minipage} &
\begin{minipage}[t]{0.2\textwidth}
\centering
\includegraphics[width=0.9\textwidth, page=3]{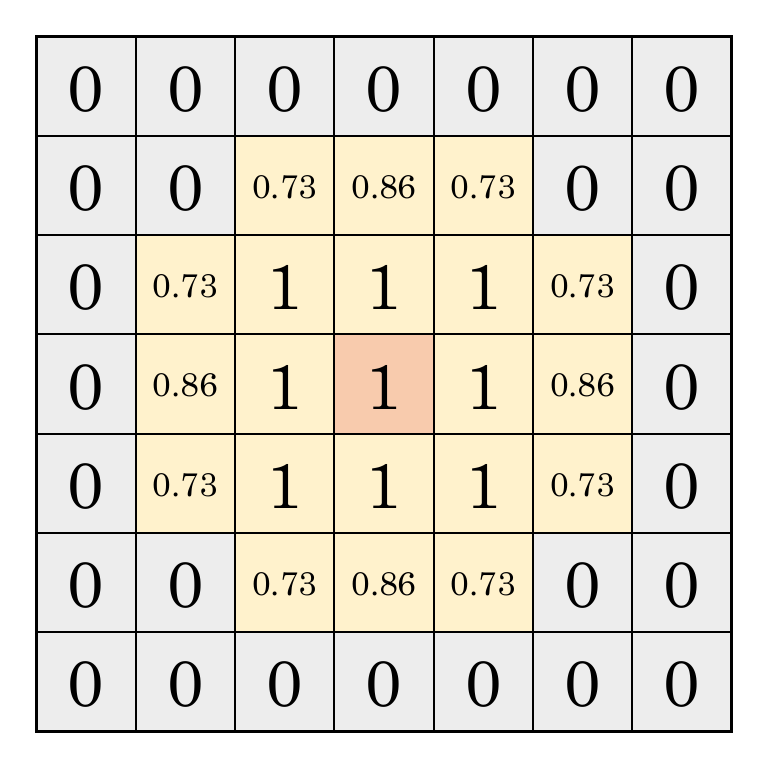}
\end{minipage}  \\
(a) & (b)
\end{tabular}
\caption{An exemplar (a) binary ground-truth (GT) map and (b) real-valued GT map.}
\label{fig:hm}
\vspace*{-3mm}
\end{wrapfigure}
\begin{equation}
y_{\mathbf{p}}^{bin} =
\begin{cases}
1 & \text{if} \;\;\; \| \mathbf{p} - \mathbf{p}^{GT} \| \leq d \\
0 & \text{otherwise},
\end{cases}
\label{eq:hm1}
\end{equation}
where $y_{\mathbf{p}}^{bin}$ is the value of the GT map at location $\mathbf{p} \in \mathbb{R}^{2}$ and $d$ is a distance threshold set differently between methods.
An exemplar binary GT map is illustrated in Figure \ref{fig:hm} (a).
The focal loss \cite{lin+2017iccv} or the combo loss \cite{taghanaki+cmig} is used to train models, all of which only supports binary maps as GT.
However, we argue that resulting prediction of existing methods tends to be less sensitive to the exact ball position, since the ball position is made obscure through the GT map generation process.
\par
To overcome this limitation, we propose a novel training scheme to make the resulting model more aware of the exact ball position.
Specifically, we first generate a {\it real-valued} GT map $\mathbf{y}^{real}$ based on the following Equation \ref{eq:hm}:
\begin{equation}
y_{\mathbf{p}}^{real} =
\begin{cases}
\min \Bigl( C \cdot \exp \bigl( - \frac{ \| \mathbf{p} - \mathbf{p}^{GT} \|^{2} }{ d^{2} } \bigr), 1 \Bigr) & \text{if} \;\;\; \| \mathbf{p} - \mathbf{p}^{GT} \| \leq d \\
0       & \text{otherwise},
\end{cases}
\label{eq:hm}
\end{equation}
where $y_{\mathbf{p}}^{real}$ is the value of the real-valued GT map at $\mathbf{p}$, while $C$ is determined so that the non-zero minimum value is set to a pre-defined value $c_{min}$.
We illustrate an exemplar real-valued GT map in Figure \ref{fig:hm} (b).
With this real-valued GT map, we optimize our model parameters by minimizing the following quality focal loss \cite{li+2020neurips,li+2020arxiv}:
\begin{equation}
L = \sum_{\mathbf{p}} \Bigl[ - | y_{\mathbf{p}} - \sigma_{\mathbf{p}} |^{\beta} \Bigl\{ ( 1-y_{\mathbf{p}} ) \log  ( 1-\sigma_{\mathbf{p}} ) + y_{\mathbf{p}} \log \sigma_{\mathbf{p}} \Bigr\} \Bigr].
\label{eq:loss}
\end{equation}
$\sigma_{\mathbf{p}}$ is the sigmoid output of the model prediction at $\mathbf{p}$ and $\beta$ is a parameter to control the down-weighting rate.
Equation \ref{eq:loss} is equivalent to the focal loss \cite{lin+2017iccv} if GT is binary.
\par
\vspace{1mm}
\begin{wrapfigure}{r}[0pt]{0.4\textwidth}
\vspace*{-4mm}
\centering
\includegraphics[width=0.4\textwidth, page=2]{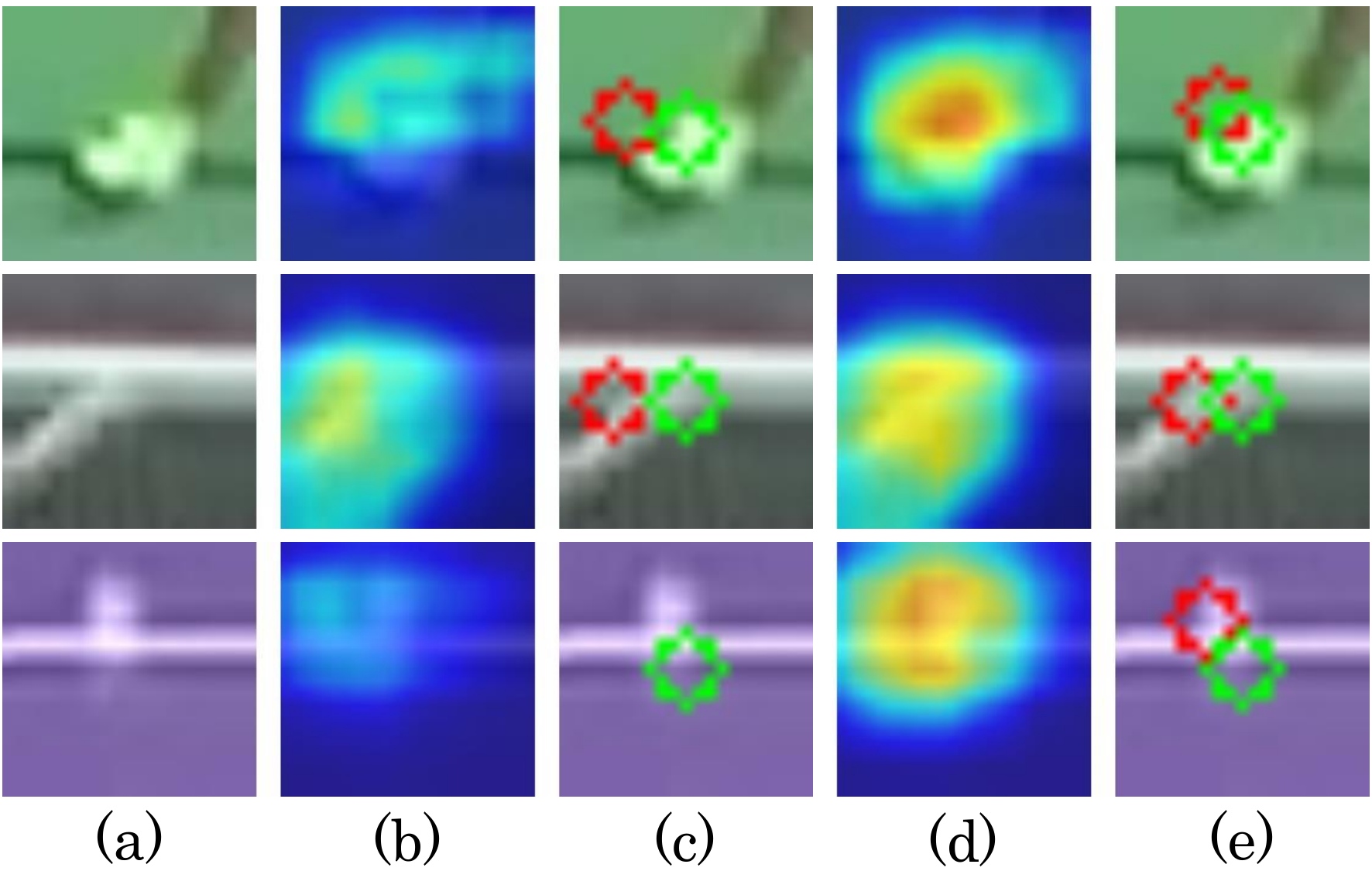}
\caption{Exemplar hard-to-localize samples found in our HLSM. In (c) and (e), a green circle represents a GT while a red one is a prediction.}
\label{fig:mining}
\end{wrapfigure}
\noindent \textbf{Hard-to-Localize Sample Mining (HLSM)}.
We empirically found that applying this position-aware GT map generation to {\it all} the training data does not statistically improve the SBDT performance.
Alternatively, we propose to apply the real-valued GT map generation scheme only to {\it hard-to-localize} samples through mining such hard examples during training.
The procedure is very simple:
After each pre-defined epoch,
we perform inference ({\it cf.} \S \ref{sec:method:infer}) with the latest model parameters over all the training sequences to find images in which predicted ball positions are far from GT positions.
For all the found images, GT are generated with Equation \ref{eq:hm}, then the model is further tuned in remaining epochs.
We show 3 hard-to-localize examples found in the above mining process in Figure \ref{fig:mining} (a).
Since their background is noisy, our model trained with {\it binary} GT maps yields blurry heatmaps as shown in (b), leading to incorrect localization or miss detection like (c).
However, through further training with {\it real-valued} GT maps, our model is able to generate clearer heatmaps as illustrated in (d), which results in more precise localization as shown in (e).
\subsection{Inference}
\label{sec:method:infer}
We first describe a baseline inference algorithm.
Given a video clip that consists of $T$ images, $N$ consecutive images are sampled in order with no overlaps (\ie, sampling step size is set to $N$), and they are preprocessed into a tensor which is fed into our trained model to produce $N$ heatmaps.
Each heatmap is binarized with a threshold $0.5$ to find connected components (\ie, blobs), and for each blob a candidate 2D ball position is estimated with its confidence.
In this baseline the ball position is computed as a geometric center and the confidence is defined as a blob size.
A ball position with the highest confidence is chosen as an inference result for each image,
while a ball is not detected if there is no blob found.
In the followings we introduce 3 simple techniques to improve this baseline inference:
\par
\vspace{1mm}
\noindent \textbf{Ball Position as a Center of Heatmap (CoH).}
We found that heatmap values in a blob can be clues to precisely estimate a ball position.
We propose to compute a ball position as the center of heatmap values, and define its confidence as a sum of heatmap values in the blob.
\par
\vspace{1mm}
\noindent \textbf{Online Tracking.}
Relying only on a detection confidence within an image could be error-prone, especially when ball-like objects appear.
We thus propose to introduce the idea of online tracking to take both detection confidence and temporal consistency into account.
Specifically, for image at $t+1$ we detect candidates using a generated heatmap, while we also predict the ball position from tracked ball positions in the previous frames.
Candidates farther from the predicted ball position than a threshold are filtered out, then a candidate with the highest confidence in the remaining candidates is selected as an inference result at $t+1$.
Following the local motion model \cite{zhou+2013icassp,zhou+2015tmm}, we compute a predicted ball position $\hat{\mathbf{p}}$ at $t+1$ as follows:
\begin{equation}
\hat{\mathbf{p}}_{t+1} = \mathbf{p}_{t} + \mathbf{v}_{t} + \frac{\mathbf{a}_{t}}{2}, \;\; \mathbf{v}_{t} = \mathbf{p}_{t} - \mathbf{p}_{t-1} + \mathbf{a}_{t}, \;\; \mathbf{a}_{t} = \mathbf{p}_{t} - 2 \mathbf{p}_{t-1} + \mathbf{p}_{t-2}.
\label{eq:pred}
\end{equation}
Notice that we exploit temporal information to just filter out inconsistent detection candidates:
Different from classical methods, we do not use filtering algorithms such as Kalman filter \cite{yu+2003icme,yu+2003icme2,yu+2003acmmm,yu+2004icip,zaveri+2004icme,kittler+2005ia,liang+2005pcm,chen+2006cesa,ren+2006eccvw,yu+2006tmm,yu+2007icme,yu+2007vcip,kim+2009cgiv,chakraborty+2013indicon} and particle filter \cite{yan+2005bmvc,abed+2006acivs,ariki+2008icme,huang+2008icpr,zhu+2008civr,beetz+2009ijcss,zhu+2009tmm}, since any performance improvement was not observed with them.
\par
\vspace{1mm}
\noindent \textbf{Oversampling.}
We also found that different MIMO sampling of the same image leads to produce diverse detection candidates.
In this work we propose to oversample the same image in different MIMO combinations, then use all the resulting candidates in the following selection step.
In \S \ref{sec:eval}, we report the results in case the step size is set to 1.
Notice that this technique may slow down inference, which is also investigated in our experiments.
\section{Dataset and Codebase}
\label{sec:dataset}
\subsection{SBDT Datasets}
\label{sec:dataset:dataset}
To evaluate the wide-applicability of SBDT algorithms, in this work we use 5 SBDT datasets from different sports categories, which are detailed in the followings.
Among them, \textcolor{blue}{Basketball} and \textcolor{blue}{Volleyball} are newly introduced datasets for SBDT, while the ground truths of \textcolor{blue}{Basketball} and \textcolor{blue}{Soccer} are newly annotated by us.
Statistics are summarized in Table \ref{tab:dataset}.
\par
\vspace{1mm}
\noindent \textbf{\textcolor{blue}{Soccer}} \cite{dorazio+2009avss}.
This dataset\footnote{\url{https://pspagnolo.jimdofree.com/download/}} was originally introduced for soccer ball and player tracking from six synchronized videos, and has been used in some SBDT works \cite{komorowski+2019visapp,komorowski+2020visapp,wang+2014cviu}.
Following \cite{komorowski+2019visapp,komorowski+2020visapp}, we use the first four video clips for training and the remaining two clips for testing.
However, we found that ball annotations provided in the original dataset are collapsed and do not localize ball position correctly.
Therefore, in this work we manually re-annotate ball position to all the frames and use the resulting annotation for training and testing.
\par
\vspace{1mm}
\noindent \textbf{Tennis} \cite{huang+2019avss}.
This dataset was introduced along with the TrackNet work \cite{huang+2019avss}, but was not used in its experiment.
Since there is no common usage for this dataset, we propose to use all the clips included in the first 7 games
as a training set, and the remainings as a testing set.
\par
\vspace{1mm}
\noindent \textbf{Badminton} \cite{sun+2020icpai}.
This dataset was introduced by the TrackNetV2 work \cite{sun+2020icpai}.
Following the dataset split defined by the authors, we use all the clips from 26 matches as a training set and the remaining 3 matches as a testing set.
\par
\noindent \textbf{\textcolor{blue}{Volleyball}}.
We introduce this dataset for the first time in the SBDT literature, by adapting video clips presented by \cite{Ibrahim+cvpr2016} and the corresponding ball annotations provided by \cite{perez+2022pr}.
We follow the manner of \cite{Ibrahim+cvpr2016} to split this dataset into training and testing sets.
Notice that in 3.7\% (178 / 4,830) of video clips any ball does not appear.
\par
\vspace{1mm}
\noindent \textbf{\textcolor{blue}{Basketball}}.
This dataset is also introduced for the first time in the SBDT literature.
We adapt the video clips provided by \cite{yan+2020eccv}, but there is no public ball annotations for this.
Therefore, we manually annotated ball positions to 45\% (81/181 games) of the whole video clips, resulting in 275,328 annotated images composed of 3,824 video clips.
Currently, this is the largest SBDT dataset.
Notice that the average ball displacement between consecutive frames is the largest among the five datasets
({\it cf.} Table \ref{tab:dataset}).
Also, camera frequently moves and zooms in rapidly to follow where play happens, which causes a complex ball trajectory in a video.
\begin{table}[t]
\centering
\scalebox{0.73}{
\begin{tabular}{@{}l|c|c|cccc|cccc@{}}
\toprule
                                                & &     & \multicolumn{4}{c|}{Train}                        & \multicolumn{4}{c}{Test}                \\
                                                & resolution & FPS & games & clips & frames & disp.[pixel]           & games & clips & frames & disp. \\ \midrule
\textcolor{blue}{Soccer} \cite{dorazio+2009avss} & $1920\times1080$ & 25 & 1 & 4 & 11994 & $10.4 \pm 10.0$ & 1 & 2 & 5999 & $15.7 \pm 13.0$ \\
Tennis \cite{huang+2019avss}                    & $1280\times720$ & 30  & 7        & 65       & 14160     & $15.3 \pm 13.0$ & 3        & 30       & 5675      & $13.6 \pm 10.2$ \\
Badminton \cite{sun+2020icpai}                  & $1280\times720$ & 30  & 26       & 172      & 78558     & $11.8 \pm 12.2$ & 3        & 29       & 12656     & $12.5 \pm 12.9$ \\
\textcolor{blue}{Volleyball} & $1280\times720$ & N/A & 39 & 3493 & 143213 & $14.4 \pm 11.4$ & 16 & 1337 & 54817 & $15.1 \pm 11.5$ \\
\textcolor{blue}{Basketball} & $1920\times1080$ & N/A & 70 & 3392 & 244224 & $33.7 \pm 21.8$ & 11 & 432 & 31104 & $33.9 \pm 21.4$ \\
\bottomrule
\end{tabular}
}
\caption{Summary of 5 SBDT datasets used in our evaluation. Among them, \textcolor{blue}{Volleyball} and \textcolor{blue}{Basketball} are newly introduced in this work. Also, for \textcolor{blue}{Soccer} and \textcolor{blue}{Basketball} we provide novel frame-wise manual annotations of 2D ball position. In this table, ``resolution'' represents the majority of image resolution in the dataset and ``disp.'' represents the average ball displacement in pixel between consecutive frames. Notice that frame per second (FPS) of \textcolor{blue}{Volleyball} and \textcolor{blue}{Basketball} are unknown (\ie, N/A), since they are not provided by adapted image sequences.}
\label{tab:dataset}
\end{table}
\subsection{Codebase of Existing SBDT Methods}
\label{sec:eval:baseline}
Most existing SBDT implementations have not been made public.
While a few exceptions exist\footnote{\url{https://nol.cs.nctu.edu.tw:234/open-source/TrackNetv2}}\footnote{\url{https://nol.cs.nctu.edu.tw:234/open-source/TrackNet}}, unfortunately they are strongly tied up with particular datasets, thus difficult to be applied to others.
Therefore, here we re-implement state-of-the-art SBDT methods to perform comparison on various SBDT datasets.
In particular, we implemented \textbf{DeepBall} \cite{komorowski+2019mva,komorowski+2020visapp}, \textbf{BallSeg} \cite{zandycke+2019mmsports}, \textbf{TrackNetV2} \cite{sun+2020icpai} and \textbf{MonoTrack} \cite{liu+2022cvprw}.
For DeepBall, since its original model is very small ($< 0.1$M parameters), we built a variant by simply increasing intermediate feature dimension, which is called \textbf{DeepBall-Large} in the followings.
Also, we deployed an unpublished variant\footnote{\url{https://github.com/Chang-Chia-Chi/TrackNet-Badminton-Tracking-tensorflow2}} of TrackNetV2, where residual connection and transposed convolution are additionally employed.
We call this variant as \textbf{ResTrackNetV2}.
\par
Notice that while we basically followed the settings proposed by authors, for some methods minor modifications were made for performance improvement.
We provide these implementation details in Appendix \ref{appendix:imple}.
\par
We report the performances of our SOTA re-implementations in Table \ref{tab:bench}.
It shows that the accuracy of our TrackNetV2 \cite{sun+2020icpai} implementation on the Badminton dataset is 85.6, while Table IV in \cite{sun+2020icpai} shows that the original implementation scores 85.2, which indicates the correctness (or, superiority) of our TrackNetV2 implementation.
Unfortunately, such a validation cannot be performed for the remaining five methods:
The original DeepBall \cite{komorowski+2019visapp} was evaluated on the Soccer dataset, but its original annotation is collapsed ({\it cf}. \S \ref{sec:dataset:dataset}), which makes the validation intractable.
For BallSeg \cite{zandycke+2019mmsports}, neither its specific architecture is presented nor the benchmark is publicly available.
The MonoTrack paper \cite{liu+2022cvprw} does not explain their experimental protocol at all, and the remaining two (DeepBall-Large and ResTrackNet) are simple extensions of existing methods proposed by us, which have no reference implementations.
\begin{table*}[t]
\scalebox{0.49}{
\centering
\begin{tabular}{@{}lc||cccc|cccc|cccc|cccc|cccc@{}}
\toprule
            &           & \multicolumn{4}{c|}{Soccer} & \multicolumn{4}{c|}{Tennis} & \multicolumn{4}{c|}{Badminton} & \multicolumn{4}{c|}{Volleyball} & \multicolumn{4}{c}{Basketball} \\
            & \# param. & F1 $\uparrow$ & Acc. $\uparrow$ & AP $\uparrow$ & FPS $\uparrow$ & F1 & Acc. & AP & FPS & F1 & Acc. & AP & FPS & F1 & Acc. & AP & FPS & F1 & Acc. & AP & FPS \\ \midrule
DeepBall \cite{komorowski+2019visapp,komorowski+2019mva} & 0.1M & 44.5 & 92.7 & 26.3 & 44.6 & 47.4 & 32.3 & 47.0 & 52.1 & 52.4 & 38.6 & 60.0 & 57.1 & 64.4 & 50.7 & 49.2 & \textcolor{teal}{21.1} & 0.0 & 12.9 & 0.0 & 30.3 \\
DeepBall-Large & 1.0M & 44.9 & 89.5 & 34.0 & 42.0 & 46.7 & 31.6 & 35.1 & 47.7 & 50.6 & 36.8 & 59.5 & 53.0 & 70.4 & 57.5 & 56.5 & \textcolor{teal}{21.1} & 57.2 & 47.5 & 36.6 & 30.9 \\
BallSeg \cite{zandycke+2019mmsports} & 12.7M & 36.1 & 92.6 & 20.0 & \textcolor{teal}{64.8} & 71.7 & 57.5 & 56.8 & \textcolor{teal}{62.7} & 79.9 & 72.2 & 68.4 & 75.0 & 19.5 & 17.5 & 8.5 & 18.2 & 16.8 & 20.5 & 5.3 & 29.5 \\
TrackNetV2 \cite{sun+2020icpai} & 11.3M & \textcolor{blue}{86.6} & \textcolor{teal}{97.7} & 77.2 & \textcolor{red}{66.0} & 89.4 & 81.4 & 80.6 & 55.3 & 90.5 & 85.6 & 83.6 & \textcolor{red}{77.0} & 83.6 & 73.8 & 72.3 & 17.6 & 78.8 & 69.3 & 64.6 & 28.0 \\
ResTrackNetV2 & 1.2M & 84.6 & 97.4 & 75.5 & 56.2 & 90.3 & 82.8 & 81.7 & 59.0 & 89.4 & 84.0 & 82.2 & 71.3 & 84.2 & 74.7 & \textcolor{blue}{74.7} & \textcolor{red}{28.6} & 77.9 & 68.2 & \textcolor{blue}{66.0} & \textcolor{red}{38.2} \\
MonoTrack \cite{liu+2022cvprw} & 2.9M & 85.2 & 97.4 & \textcolor{blue}{78.6} & 58.0 & \textcolor{blue}{92.1} & \textcolor{blue}{85.9} & \textcolor{blue}{87.3} & \textcolor{red}{64.1} & \textcolor{blue}{90.9} & \textcolor{blue}{85.9} & \textcolor{blue}{84.9} & \textcolor{teal}{75.5} & \textcolor{blue}{85.1} & \textcolor{blue}{75.9} & 72.1 & 19.7 & \textcolor{teal}{80.8} & \textcolor{teal}{71.3} & 65.3 & \textcolor{teal}{32.1} \\ \midrule
\rowcolor[gray]{0.9}
WASB (Ours, Step=3) & 1.5M & \textcolor{red}{88.3} & \textcolor{red}{97.9} & \textcolor{teal}{83.6} & 55.7 & \textcolor{teal}{94.0} & \textcolor{teal}{89.0} & \textcolor{teal}{91.0} & 58.2 & \textcolor{teal}{91.6} & \textcolor{teal}{87.0} & \textcolor{teal}{88.5} & 70.4 & \textcolor{teal}{86.5} & \textcolor{teal}{77.9} & \textcolor{teal}{79.9} & 18.0 & 80.6 & \textcolor{teal}{71.3} & \textcolor{teal}{71.5} & 30.2 \\
\rowcolor[gray]{0.9}
WASB (Ours, Step=1) & 1.5M & \textcolor{teal}{88.2} & \textcolor{red}{97.9} & \textcolor{red}{86.2} & 23.6 & \textcolor{red}{95.6} & \textcolor{red}{91.8} & \textcolor{red}{94.2} & 35.2 & \textcolor{red}{93.1} & \textcolor{red}{89.0} & \textcolor{red}{91.6} & 34.3 & \textcolor{red}{88.0} & \textcolor{red}{80.0} & \textcolor{red}{83.2} & 15.8 & \textcolor{red}{82.6} & \textcolor{red}{73.4} & \textcolor{red}{77.1} & 22.3 \\ \bottomrule
\end{tabular}
}
\caption{Benchmark results of SBDT methods on 5 SBDT datasets. We set the distance threshold $\tau=4$ [pixel] to compute F1, Accuracy (Acc.) and Average Precision (AP), all of which are shown as percentages. Red values are the best while green values are the second-best among all the methods. Blue values are the best in existing methods.}
\label{tab:bench}
\end{table*}
\section{EVALUATION}
\label{sec:eval}
Here we report quantitative evaluations of our proposed method, WASB, using the datasets and codebases established in \S \ref{sec:dataset}.
Qualitative results are presented in Appendix \ref{appendix:result}.
\subsection{Evaluation Metrics}
\label{sec:eval:metric}
We evaluate SBDT models using F1, Accuracy (Acc.) and Average Precision (AP).
With a distance threshold $\tau$ [pixel], for each frame we calculate the distance between a predicted ball position and a ground truth to classify the prediction into true positive, true negative, false positive or false negative.
F1 and Acc. can be directly computed with the results, while AP is computed over all the positive results with prediction confidences.
\subsection{Implementation Details}
\label{sec:eval:imple}
Following TrackNetV2 and its variants \cite{sun+2020icpai,liu+2022cvprw}, $N$ ({\it cf}. \S \ref{sec:method:cnn}) is set to 3 and each image is resized to $288 \times 512$ to be fed into our model.
We train our model from scratch with Adam optimizer \cite{kingma+2015iclr} for 30 epochs.
The batch size is set to 8 for both training and testing.
To generate GT maps, $d$ ({\it cf}. \S \ref{sec:method:train}) is set to 2.5 while $c_{min}$ is set to 0.7.
We run HLSM ({\it cf}. \S \ref{sec:method:train}) at the beginning of epoch 20, while we didn't observe performance improvement with more trials.
We performed all the following experiments on an Ubuntu server with 4 V100 GPUs.
\begin{figure*}[t]
\centering
\includegraphics[width=1.0\textwidth, page=1]{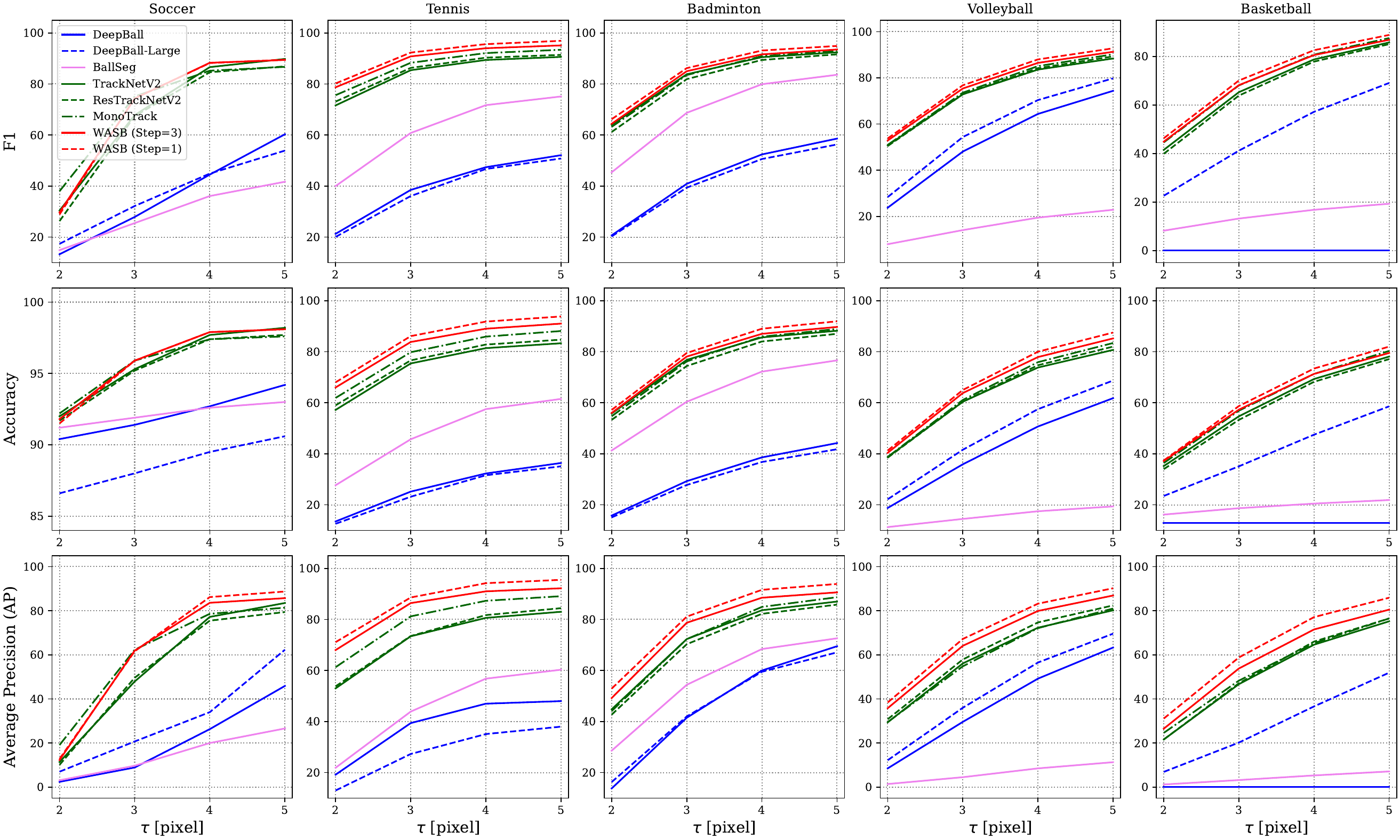}
\caption{F1 (first row), Accuracy (second row) and Average Precision (third row) of SBDT methods with different distance threshold $\tau$ [pixel] on 5 SBDT datasets.}
\label{fig:benchmark2}
\end{figure*}
\subsection{Main Results}
\label{sec:eval:comp}
Table \ref{tab:bench} shows the benchmark results of SBDT methods on our datasets, using the fixed distance threshold $\tau=4$ [pixel].
For our proposed WASB, we show the results where the step size is set to 3 (\ie, no oversampling) and 1 ({\it cf.} \S \ref{sec:method:infer}).
We can clearly see that WASB results dominate the best and the second-best SBDT performance over the most metrics in sports categories covered by our datasets.
Also, with respect to AP, our best models significantly outperform the best existing methods by 7.8 \textasciitilde 16.8 \%.
Notice that WASB is not the fastest among the methods.
However, it can still be processed over 30 FPS on 4 out of 5 datasets, which is reasonable efficiency for real-time inference.
\par
Figure \ref{fig:benchmark2} shows F1, Accuracy and AP scores of SBDT methods with different distance thresholds.
Interestingly, the performances of DeepBall \cite{komorowski+2019mva,komorowski+2020visapp} and BallSeg \cite{zandycke+2019mmsports} heavily depend on the dataset,
while TrackNetV2 \cite{sun+2020icpai}, ResTrackNetV2 and MonoTrack \cite{liu+2022cvprw} stably yield good results through the datasets.
Compared to these methods, WASB consistently achieves higher performance with most of the threshold settings on all the sports categories, which indicates the wide-applicability of our approach.
\subsection{Ablation Studies}
\label{sec:eval:ablation}
Table \ref{tab:ablation:model} shows the ablation results with respect to the model design discussed in \S \ref{sec:method:cnn}.
As expected, removing strides can contribute to improving the model performance through the datasets.
Also as anticipated, removing strides from the stem seems to slow down inference.
However, the actual impact is not so severe, and in some cases (\eg, volleyball) we do not observe the degradation of efficiency.
\par
Table \ref{tab:ablation:misc} represents the ablation results to evaluate the techniques introduced in \S \ref{sec:method:train} and \S \ref{sec:method:infer}.
We can see that each technique complementarily ameliorate the SBDT performance with a few exceptions (\eg, online tracking does not contribute on the Soccer and Badminton datasets).
Interestingly, {\it even without any techniques}, our method is superior to the best of existing methods ({\it cf.} first row in Table \ref{tab:ablation:misc}).
This indicates the superiority of our high-resolution feature extraction model to existing approaches.
\subsection{Limitation}
\label{sec:eval:limitation}
As with the most of existing SBDT methods, our method, WASB, assumes a ball-game video as an input, and predicts at most one ball location (\ie, a $(x,y)$-coordinate) for each frame.
Therefore, one apparent limitation is that WASB cannot be applied to sports in which multiple balls are used simultaneously
(\eg, billiards \cite{rea+2004ivr}).
Our method can be applied to videos both captured by fixed cameras and including camera motion, which is validated with our Basketball dataset ({\it cf}. \S \ref{sec:dataset:dataset}).
While there are no theoretical limitations with respect to frame resolution and frame rate, our validation is limited to standard frame resolutions (\eg, HD, FHD) and frame rates (\eg, 25-30 FPS).
\section{CONCLUSION}
\label{sec:conc}
In this paper we proposed a Widely Applicable Strong Baseline (WASB) for Sports Ball Detection and Tracking (SBDT).
Extensive experiments on 5 SBDT datasets from different sports categories demonstrate that our WASB achieves substantially better performance than 6 state-of-the-art (SOTA) SBDT methods on all the datasets.
We achieve this by introducing two novel SBDT datasets, providing two new manual annotations, and re-implementing all the SOTA methods.
In the future research, we explore to make our baseline more efficient while keeping its performance.
Extending SBDT datasets (\eg, dataset scale, sports category) is also an interesting research direction.
\begin{table*}[t]
\centering
\scalebox{0.51}{
\begin{tabular}{@{}lc||cccc|cccc|cccc|cccc|cccc@{}}
\toprule
            &           & \multicolumn{4}{c|}{Soccer} & \multicolumn{4}{c|}{Tennis} & \multicolumn{4}{c|}{Badminton} & \multicolumn{4}{c|}{Volleyball} & \multicolumn{4}{c}{Basketball} \\
            & \# param. & F1 $\uparrow$ & Acc. $\uparrow$ & AP $\uparrow$ & FPS $\uparrow$ & F1 & Acc. & AP & FPS & F1 & Acc. & AP & FPS & F1 & Acc. & AP & FPS & F1 & Acc. & AP & FPS \\ \midrule
Figure \ref{fig:arch} (a) & 1.5M & 81.7 & 96.9 & 71.7 & 85.7 & 85.5 & 75.4 & 75.6 & 56.7 & 86.8 & 80.4 & 80.3 & 77.1 & 84.3 & 74.7 & 77.0 & 17.6 & 77.4 & 67.3 & 67.1 & 30.8  \\
Figure \ref{fig:arch} (b) & 1.5M & 86.4 & 97.6 & 79.0 & 76.7 & 91.9 & 85.4 & 86.7 & 60.3 & 90.5 & 85.5 & 86.2 & 76.2 & 85.0 & 75.8 & 77.2 & 17.9 & 80.4 & 71.0 & 71.4 & 28.7 \\
Figure \ref{fig:arch} (c) & 1.5M & 88.3 & 97.9 & 83.6 & 55.7 & 94.0 & 89.0 & 91.0 & 58.2 & 91.6 & 87.0 & 88.5 & 70.4 & 86.5 & 77.9 & 79.9 & 18.0 & 80.6 & 71.3 & 71.5 & 30.2 \\ \bottomrule
\end{tabular}
}
\caption{Ablations with respect to the model design ({\it cf.} \S \ref{sec:method:train}). Notice that in all the cases we do not adapt oversampling ({\it cf.} \S \ref{sec:method:infer}) for inference. }
\label{tab:ablation:model}
\end{table*}
\begin{table*}[t]
\centering
\scalebox{0.6}{
\begin{tabular}{@{}cccc||ccc|ccc|ccc@{}}
\toprule
 & & & & \multicolumn{3}{c|}{Soccer} & \multicolumn{3}{c|}{Tennis} & \multicolumn{3}{c}{Badminton} \\
HLSM (\S \ref{sec:method:train}) & CoH (\S \ref{sec:method:infer}) & Online Tracking (\S \ref{sec:method:infer}) & Step=1 (\S \ref{sec:method:infer}) & F1  $\uparrow$ & Acc. $\uparrow$       & AP $\uparrow$ & F1 & Acc. & AP & F1 & Acc. & AP \\ \midrule
\multicolumn{4}{c||}{( The best scores of existing methods ({\it cf.} Table \ref{tab:bench})) } & 85.2 & 97.7 & 78.6 & 92.1 & 85.9 & 87.3 & 90.9 & 85.9 & 84.9 \\ \midrule
           &            &             &            & 87.3 & 97.7 & 80.1 & 93.1 & 88.1 & 88.5 & 91.1 & 86.3 & 85.5 \\
\Checkmark &            &             &            & 87.8 & 97.8 & 81.1 & 93.7 & 88.6 & 89.4 & 91.4 & 86.6 & 86.2 \\
\Checkmark & \Checkmark &             &            & 88.3 & 97.9 & 83.6 & 93.9 & 88.8 & 90.8 & 91.6 & 87.0 & 88.5 \\
\Checkmark & \Checkmark & \Checkmark  &            & 88.3 & 97.9 & 83.6 & 94.0 & 89.0 & 91.0 & 91.6 & 87.0 & 88.5 \\
\Checkmark & \Checkmark & \Checkmark  & \Checkmark & 88.2 & 97.9 & 86.2 & 95.6 & 91.8 & 94.2 & 93.1 & 89.0 & 91.6 \\ \bottomrule
\end{tabular}
}
\caption{Ablation results with respect to our proposed training ({\it cf.} \S \ref{sec:method:train}) and inference ({\it cf.} \S \ref{sec:method:infer}) schemes on the Soccer, Tennis and Badminton datasets.}
\label{tab:ablation:misc}
\end{table*}

\appendix
\section{Details of Existing SBDT Methods}
\label{appendix:imple}
As is mentioned in \S 4.2, we re-implemented 6 state-of-the-art (SOTA) sports ball detection and tracking (SBDT) algorithms in our codebase, 4 of which have been proposed in the recent literature \cite{komorowski+2019mva,komorowski+2020visapp,zandycke+2019mmsports,sun+2020icpai,liu+2022cvprw} and the remaining 2 of which are their variants.
We basically followed the default implementation settings proposed by authors, meanwhile we found that their performance can be boosted by simple modifications.
In the following we describe the details of SOTA SBDT methods including modifications made by us.
\par
\vspace{1mm}
\noindent \textbf{DeepBall} \cite{komorowski+2019mva,komorowski+2020visapp}. This is a small convolutional neural network (CNN) that is originally proposed to detect a soccer ball.
Unfortunately, its official implementation has not been publicly available.
DeepBall takes a single frame to produce the heatmap representing ball position via aggregating multi-scale intermediate feature maps.
At inference time, a ball position is determined by simply detecting a peak from the heatmap.
Model training is performed by minimizing the pixel cross-entropy (CE) loss between model predictions and ground truth (GT) binary maps.
The GT binary map is produced by setting a true ball position and its nearest neighbours as foreground.
Adam optimizer \cite{kingma+2015iclr} is used to train the model, and hard negative mining \cite{liu+2016eccv} is employed to mitigate the effect of foreground-background class imbalance.
Notice that we directly followed the above settings for our re-implementation.
\par
\vspace{1mm}
\noindent \textbf{DeepBall-Large}.
Through the re-implementation of DeepBall, we found that the original model is too small ($< 0.1$M parameters) to be applied to other ball-game datasets ({\it cf.} Table 2 in our main body).
To increase the model capacity, we made the following two modifications to the original DeepBall model: (1) The depths of block \{1, 2, 3\} are increased from \{8, 16, 32\} to \{48, 96, 192\}, (2) a kernel size of the stem is set to 3.
Here we call the resulting variant of DeepBall as DeepBall-Large.
Its model training is the same with the original.
\par
\vspace{1mm}
\noindent \textbf{BallSeg} \cite{zandycke+2019mmsports}.
This is a variant of ICNet \cite{zhao+2018eccv} originally proposed to detect a basketball.
Its official implementation has not been publicly available.
BallSeg takes two consecutive frames by concatenating a frame of interest with its difference to another frame.
The model is trained using the Stochastic Gradient Descent (SGD) applied on the pixel-wise CE loss.
Since the specific ICNet architecture used to build BallSeg is not described in the original paper, we chose to adapt the smallest model provided in the official ICNet repository\footnote{\url{https://github.com/hszhao/ICNet}}.
Also, we found that model training is failed when the proposed loss and optimizer are used.
Instead, we employed the focal loss \cite{lin+2017iccv} and Adam optimizer \cite{kingma+2015iclr} to successfully train BallSeg, then evaluated the performance of resulting models in our experiments ({\it cf.} \S 4 in our manuscript).
\par
\vspace{1mm}
\noindent \textbf{TrackNetV2} \cite{sun+2020icpai}.
This is a UNet-based \cite{ronneberger+2015miccai} SBDT model originally proposed to detect a shuttlecock from badminton videos.
The authors proposed multiple-in multiple-out (MIMO) design to efficiently capture ball dynamics:
Three consecutive frames are concatenated along the channel dimension, then the resulting tensor is fed into the model that generates corresponding three heatmaps.
The model is trained using the Adadelta \cite{zeiler2012arxiv} optimizer applied on the the focal loss \cite{lin+2017iccv}.
Though its official implementation has been public\footnote{\url{https://nol.cs.nctu.edu.tw:234/open-source/TrackNetv2}}, unfortunately it is strongly tied up with the badminton dataset thus is difficult to adapt to other sports datasets.
Therefore, we re-implemented TrackNetV2 following the above settings while being applicable it to various sports datasets.
\par
\vspace{1mm}
\noindent \textbf{ResTrackNetV2}.
We found that there is a public SBDT repository\footnote{\url{https://github.com/Chang-Chia-Chi/TrackNet-Badminton-Tracking-tensorflow2}} that extends TrackNet \cite{huang+2019avss} by introducing residual connections \cite{he+2016cvpr}.
Based on this idea, we also added a residual connection to each encoder/decoder block in TrackNetV2 \cite{sun+2020icpai} to promote the model training.
Also, we decreased the channel dimension of encoder/decoder blocks, which results in almost one-tenth model parameters compared to the original TrackNetV2.
Here we call this variant as ResTrackNetV2.
We trained this model with the same manner with TrackNetV2.
\par
\vspace{1mm}
\noindent \textbf{MonoTrack} \cite{liu+2022cvprw} is another variant of TrackNetV2 \cite{sun+2020icpai}, which removes some convolution layers while adding skip connections.
One notable difference from TrackNetV2 is that they adopt the combo loss \cite{taghanaki+cmig} in model training.
Since its official implementation has not been publicly available, we also re-implemented this method following settings described in \cite{liu+2022cvprw}.
\section{Qualitative Results and Error Analysis}
\label{appendix:result}
\begin{figure}[htbp]
\begin{tabular}{c}
\begin{minipage}[t]{1.0\hsize}
\centering
\includegraphics[keepaspectratio, scale=0.25, page=2]{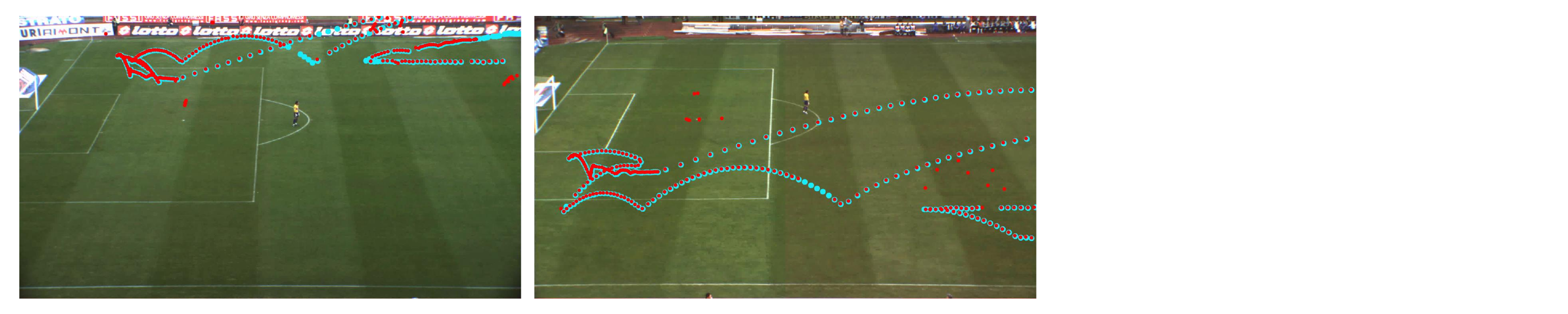}
\end{minipage}
\\
(a) Soccer
\\
\begin{minipage}[t]{1.0\hsize}
\centering
\includegraphics[keepaspectratio, scale=0.25, page=3]{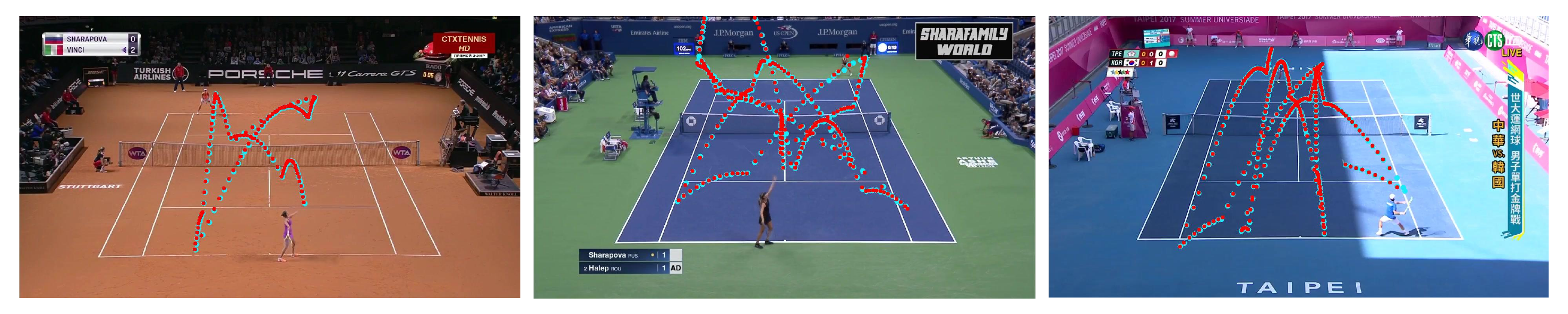}
\end{minipage}
\\
(b) Tennis
\\
\begin{minipage}[t]{1.0\hsize}
\centering
\includegraphics[keepaspectratio, scale=0.25, page=4]{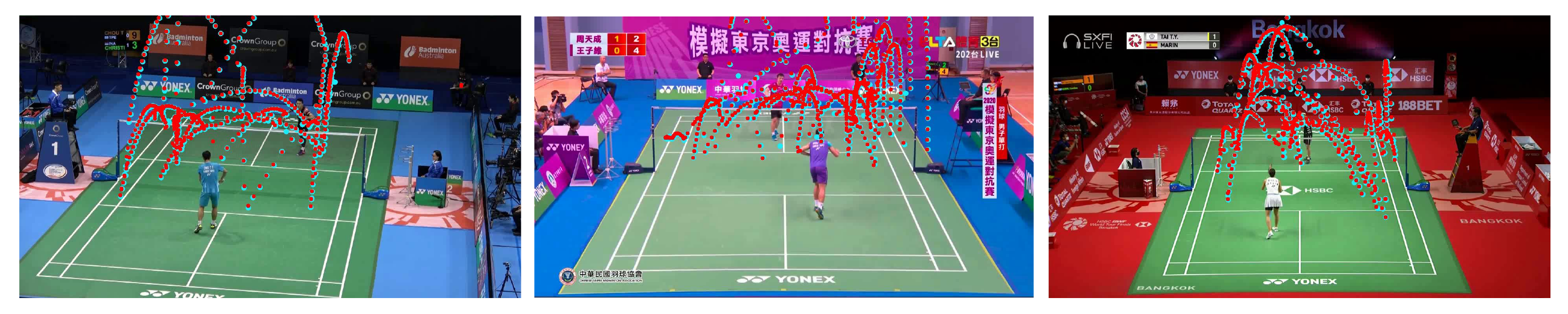}
\end{minipage}
\\
(c) Badminton
\\
\begin{minipage}[t]{1.0\hsize}
\centering
\includegraphics[keepaspectratio, scale=0.25, page=5]{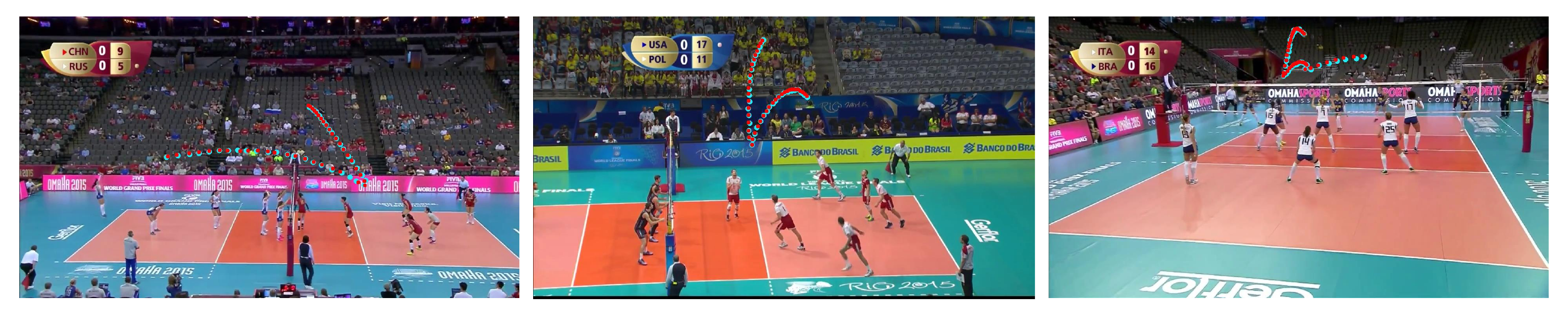}
\end{minipage}
\\
(d) Volleyball
\\
\begin{minipage}[t]{1.0\hsize}
\centering
\includegraphics[keepaspectratio, scale=0.25, page=6]{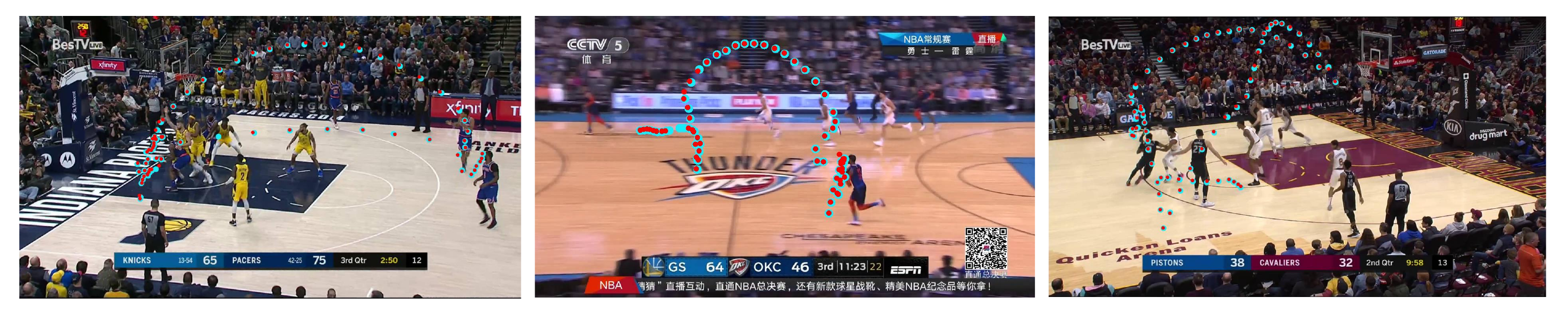}
\end{minipage}
\\
(e) Basketball
\end{tabular}
\caption{Exemplar qualitative results of our proposed method on each sports category in our dataset collection. A red circle represents a detection result while a light blue circle represents a ground truth ball position. The ball trajectory is overlaid on the first frame in each video clip. Best viewed in color.}
\label{fig:qualitative}
\end{figure}

Figure \ref{fig:qualitative} shows typical SBDT results of our proposed method, WASB ({\it cf.} \S 3 in our manuscript).
These results demonstrates that WASB correctly track balls from video clips of different sports categories.
Interestingly, we can see that sports balls can be tracked from video clips with very different viewpoints (\eg, (d) Volleyball), and also from video clips including fast camera motion (\eg, (e) Basketball).
\par
\begin{figure}[htbp]
\begin{tabular}{c}
\begin{minipage}[t]{1.0\hsize}
\centering
\includegraphics[keepaspectratio, scale=0.25, page=2]{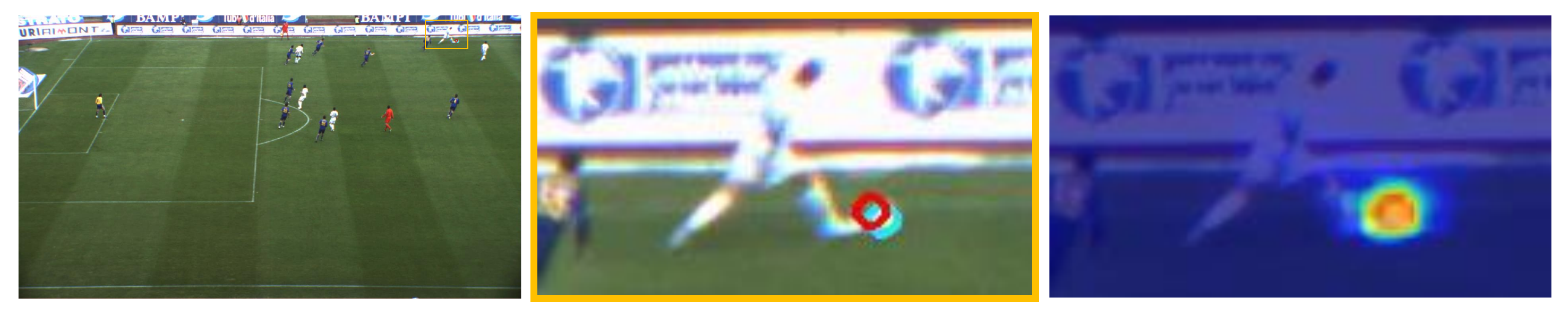}
\end{minipage}
\\
(a) Soccer
\\
\begin{minipage}[t]{1.0\hsize}
\centering
\includegraphics[keepaspectratio, scale=0.25, page=3]{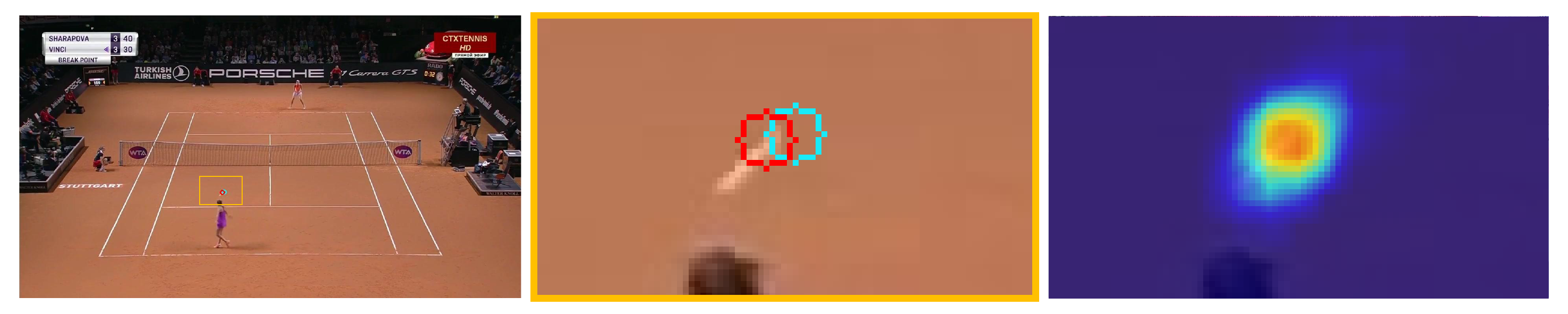}
\end{minipage}
\\
(b) Tennis
\\
\begin{minipage}[t]{1.0\hsize}
\centering
\includegraphics[keepaspectratio, scale=0.25, page=4]{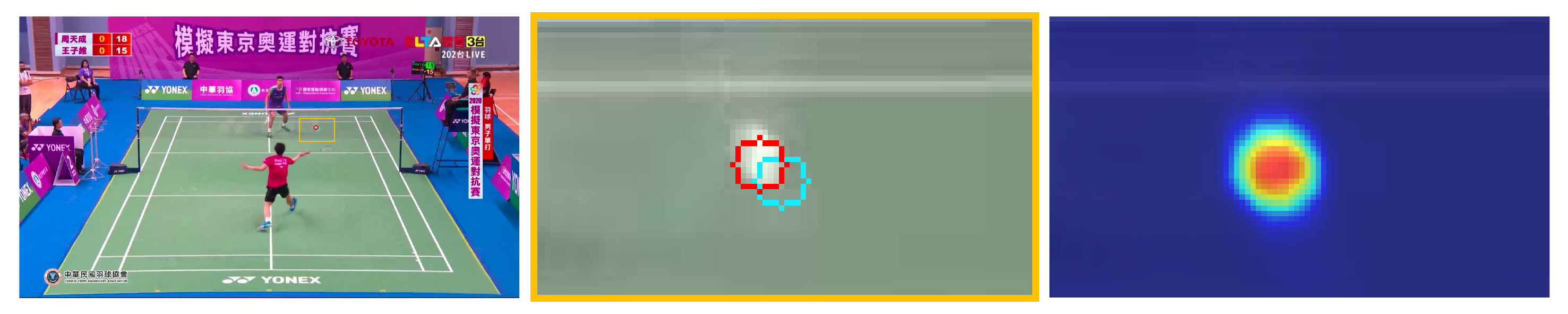}
\end{minipage}
\\
(c) Badminton
\\
\begin{minipage}[t]{1.0\hsize}
\centering
\includegraphics[keepaspectratio, scale=0.25, page=5]{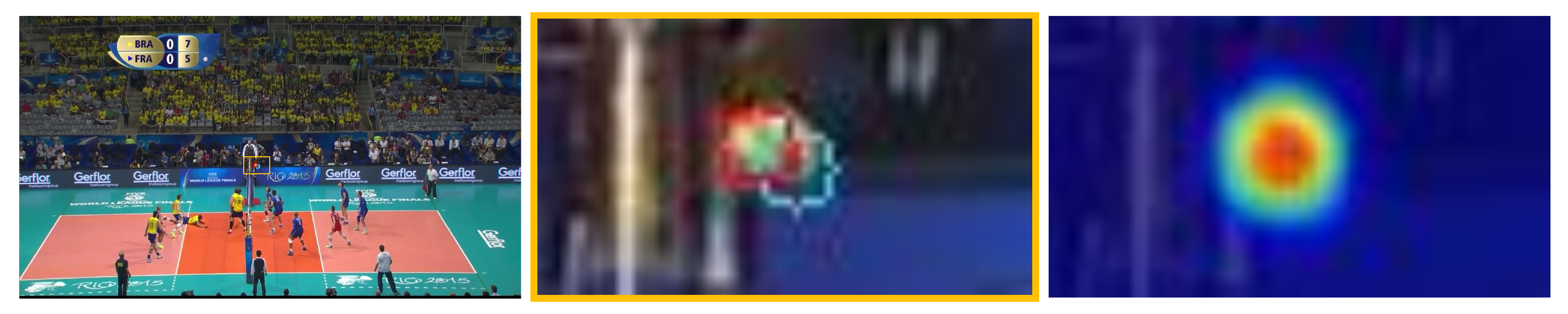}
\end{minipage}
\\
(d) Volleyball
\\
\begin{minipage}[t]{1.0\hsize}
\centering
\includegraphics[keepaspectratio, scale=0.25, page=6]{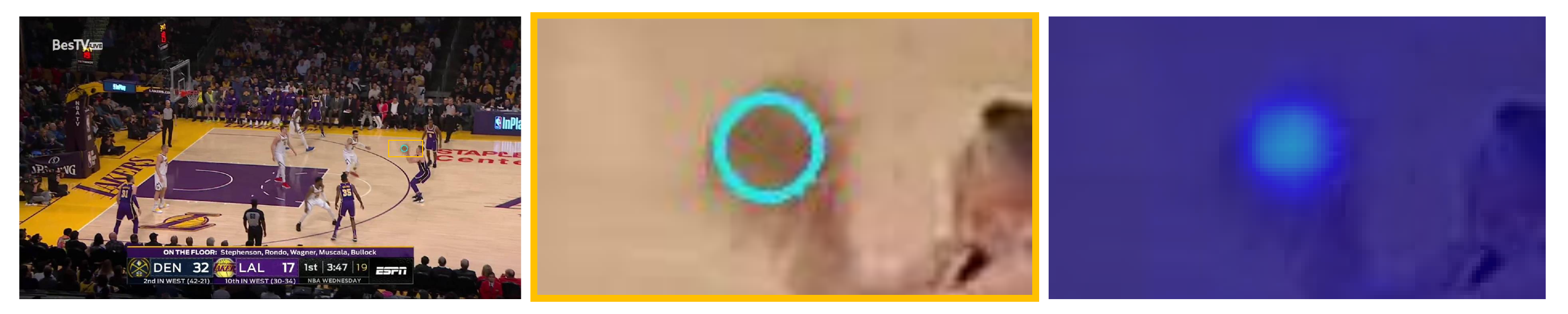}
\end{minipage}
\\
(e) Basketball
\end{tabular}
\caption{Exemplar error modes of our proposed method. A red circle represents a detection result while a light blue circle represents a ground truth ball position. Results in the second column is the zoom of yellow rectangle areas in the first column, and the third column shows the corresponding heatmaps produced by our model. Best viewed in color.}
\label{fig:error}
\end{figure}

Figure \ref{fig:error} shows some error modes of our proposed method.
For example, the result (a) (\ie, Soccer) represents a false positive, while the result (e) (\ie, Basketball) shows a false negative.
We can see that in (a) the model detection is not precisely aligned due to the noisy background (\eg, player shoes), while in (e) a ball cannot be detected because it is blurry and ambiguous.
The results (b), (c) and (d) (\ie, Tennis, Badminton, Volleyball) also represent false positives.
Interestingly, however, in these examples model detections (red circles) seem to capture true ball positions (light blue) more correctly than manually annotated ground truths.
There results indicate a potential of WASB surpassing human ball localization performance.

\newpage

\bibliography{refs}

\begin{thebibliography}{113}
\providecommand{\natexlab}[1]{#1}
\providecommand{\url}[1]{\texttt{#1}}
\expandafter\ifx\csname urlstyle\endcsname\relax
  \providecommand{\doi}[1]{doi: #1}\else
  \providecommand{\doi}{doi: \begingroup \urlstyle{rm}\Url}\fi

\bibitem[Almajai et~al.(2012)Almajai, Yan, de~Campos, Khan, Christmas, Windridge, and Kittler]{almajai+2012dir}
Ibrahim Almajai, Fei Yan, Teofilo de~Campos, Aftab Khan, William~J. Christmas, David Windridge, and Josef Kittler.
\newblock {Anomaly Detection and Knowledge Transfer in Automatic Sports Video Annotation}.
\newblock In \emph{Detection and Identification of Rare Audiovisual Cues}, 2012.

\bibitem[Archana and Geetha(2015)]{archana+2015pcs}
Maruthavanan Archana and M.~Kalaisevi Geetha.
\newblock {Object Detection and Tracking Based on Trajectory in Broadcast Tennis Video}.
\newblock \emph{Procedia Computer Science}, 2015.

\bibitem[Ariki et~al.(2008)Ariki, Takiguchi, and Yano]{ariki+2008icme}
Yasuo Ariki, Tetsuya Takiguchi, and Kazuki Yano.
\newblock {Digital Camera Work for Soccer Video Production with Event Recognition and Accurate Ball Tracking by Switching Search Method}.
\newblock In \emph{2008 IEEE International Conference on Multimedia and Expo}, 2008.

\bibitem[Beetz et~al.(2009)Beetz, von Hoyningen-Huene, Kirchlechner, Gedikli, Siles, Durus, and Lames]{beetz+2009ijcss}
Michael Beetz, Nico von Hoyningen-Huene, Bernhard Kirchlechner, Suat Gedikli, Francisco Siles, Murat Durus, and Martin Lames.
\newblock {ASPOGAMO: Automated Sports Games Analysis Models}.
\newblock \emph{Int. J. Comput. Sci. Sport}, 2009.

\bibitem[Chakraborty and Meher(2011)]{chakraborty+2011sic}
Bodhisattwa Chakraborty and Sukadev Meher.
\newblock {2D Trajectory-Based Position Estimation and Tracking of a Ball in a Basketball Video}.
\newblock In \emph{Second International Conference on Trends in Optics and Photonics}, 2011.

\bibitem[Chakraborty and Meher(2012)]{chakraborty+2012icspcc}
Bodhisattwa Chakraborty and Sukadev Meher.
\newblock {Real-time Position Estimation and Tracking of a Basketball}.
\newblock In \emph{2012 IEEE International Conference on Signal Processing, Computing and Control}, 2012.

\bibitem[Chakraborty and Meher(2013{\natexlab{a}})]{chakraborty+2013indicon}
Bodhisattwa Chakraborty and Sukadev Meher.
\newblock {A Trajectory-based Ball Detection and Tracking System with Applications to Shooting Angle and Velocity Estimation in Basketball Videos}.
\newblock In \emph{2013 Annual IEEE India Conference (INDICON)}, 2013{\natexlab{a}}.

\bibitem[Chakraborty and Meher(2013{\natexlab{b}})]{chakraborty+2013jo}
Bodhisattwa Chakraborty and Sukadev Meher.
\newblock {A Real-time Trajectory-based Ball Detection-and-tracking Framework for Basketball Video}.
\newblock \emph{Journal of Optics}, 2013{\natexlab{b}}.

\bibitem[Chen and Wang(2007)]{chen+2007tst}
Bingqi Chen and Zhiqiang Wang.
\newblock {A Statistical Method for Analysis of Technical Data of a Badminton Match Based on 2-D Seriate Images}.
\newblock \emph{Tsinghua Science and Technology}, 2007.

\bibitem[Chen et~al.(2007)Chen, Chen, and Lee]{chen+2007icassp}
Hua-Tsung Chen, Hsuan-Shen Chen, and Suh-Yin Lee.
\newblock {Physics-Based Ball Tracking in Volleyball Videos with its Applications to Set Type Recognition and Action Detection}.
\newblock In \emph{2007 IEEE International Conference on Acoustics, Speech and Signal Processing - ICASSP '07}, 2007.

\bibitem[Chen et~al.(2009)Chen, Tien, Chen, Tsai, and Lee]{chen+2009jvcir}
Hua-Tsung Chen, Ming-Chun Tien, Yi-Wen Chen, Wen-Jiin Tsai, and Suh-Yin Lee.
\newblock {Physics-based Ball Tracking and 3D Trajectory Reconstruction with Applications to Shooting Location Estimation in Basketball Video}.
\newblock \emph{Journal of Visual Communication and Image Representation}, 2009.

\bibitem[Chen et~al.(2012)Chen, Tsai, Lee, and Yu]{chen2012mta}
Hua-Tsung Chen, Wen-Jiin Tsai, Suh-Yin Lee, and Jen-Yu Yu.
\newblock {Ball Tracking and 3D Trajectory Approximation with Applications to Tactics Analysis from Single-camera Volleyball Sequences}.
\newblock \emph{Multimedia Tools and Applications}, 2012.

\bibitem[Chen and Zhang(2006)]{chen+2006cesa}
Wei Chen and Yu-Jin Zhang.
\newblock {Tracking Ball and Players with Applications to Highlight Ranking of Broadcasting Table Tennis Video}.
\newblock In \emph{The Proceedings of the Multiconference on "Computational Engineering in Systems Applications"}, 2006.

\bibitem[Cheng et~al.(2015)Cheng, Zhuang, Wang, Honda, and Ikenaga]{cheng+2015pcm}
Xina Cheng, Xizhou Zhuang, Yuan Wang, Masaaki Honda, and Takeshi Ikenaga.
\newblock {Particle Filter with Ball Size Adaptive Tracking Window and Ball Feature Likelihood Model for Ball's 3D Position Tracking in Volleyball Analysis}.
\newblock In \emph{Advances in Multimedia Information Processing -- PCM 2015}, 2015.

\bibitem[Cheng et~al.(2016)Cheng, Honda, Ikoma, and Ikenaga]{cheng+2016icassp}
Xina Cheng, Masaaki Honda, Norikazu Ikoma, and Takeshi Ikenaga.
\newblock {Anti-occlusion Observation Model and Automatic Recovery for Multi-view Ball Tracking in Sports Analysis}.
\newblock In \emph{2016 IEEE International Conference on Acoustics, Speech and Signal Processing (ICASSP)}, 2016.

\bibitem[Choi and Seo(2004)]{choi+2004smvp}
Kyuhyoung Choi and Yongdeuk Seo.
\newblock {Probabilistic Tracking of the Soccer Ball}.
\newblock In \emph{Statistical Methods in Video Processing}, 2004.

\bibitem[Choi and Seo(2005)]{choi+2005iciap}
Kyuhyoung Choi and Yongduek Seo.
\newblock {Tracking Soccer Ball in TV Broadcast Video}.
\newblock In \emph{Image Analysis and Processing -- ICIAP 2005}, 2005.

\bibitem[Desai et~al.(2005)Desai, Merchant, Zaveri, Ajishna, Purohit, and Phanish]{desai+2005prmi}
Uday~B. Desai, Shabbir~N. Merchant, Mukesh Zaveri, G.~Ajishna, Manoj Purohit, and H.~S. Phanish.
\newblock {Small Object Detection and Tracking: Algorithm, Analysis and Application}.
\newblock In \emph{Pattern Recognition and Machine Intelligence}, 2005.

\bibitem[D'Orazio et~al.(2009{\natexlab{a}})D'Orazio, Leo, Mosca, Spagnolo, and Mazzeo]{dorazio+2009avss}
Tiziana D'Orazio, Marco Leo, Nicola Mosca, Paolo Spagnolo, and Pier~Luigi Mazzeo.
\newblock {A Semi-automatic System for Ground Truth Generation of Soccer Video Sequences}.
\newblock In \emph{2009 Sixth IEEE International Conference on Advanced Video and Signal Based Surveillance}, 2009{\natexlab{a}}.

\bibitem[D'Orazio et~al.(2009{\natexlab{b}})D'Orazio, Leo, Spagnolo, Mazzeo, Mosca, Nitti, and Distante]{dorazio+2009tcsvt}
Tiziana D'Orazio, Marco Leo, Paolo Spagnolo, Pier~Luigi Mazzeo, Nicola Mosca, Massimiliano Nitti, and Arcangelo Distante.
\newblock {An Investigation Into the Feasibility of Real-Time Soccer Offside Detection From a Multiple Camera System}.
\newblock \emph{IEEE Transactions on Circuits and Systems for Video Technology}, 2009{\natexlab{b}}.

\bibitem[D’Orazio et~al.(2009)D’Orazio, Leo, Spagnolo, Nitti, Mosca, and Distante]{dorazio+2009cviu}
Tiziana D’Orazio, Marco Leo, Paolo Spagnolo, Massimiliano Nitti, Nicola Mosca, and Arcangelo Distante.
\newblock {A Visual System for Real Time Detection of Goal Events during Soccer Matches}.
\newblock \emph{Computer Vision and Image Understanding}, 2009.

\bibitem[Ekinci and Gokmen(2008)]{ekinci+2008vis}
Baris~David Ekinci and Muhittin Gokmen.
\newblock {A Ball Tracking System for Offline Tennis Videos}.
\newblock In \emph{Proceedings of the 1st WSEAS International Conference on Visualization, Imaging and Simulation}, 2008.

\bibitem[El~Abed et~al.(2006)El~Abed, Dubuisson, and B{\'e}r{\'e}ziat]{abed+2006acivs}
Abir El~Abed, S{\'e}verine Dubuisson, and Dominique B{\'e}r{\'e}ziat.
\newblock {Comparison of Statistical and Shape-Based Approaches for Non-rigid Motion Tracking with Missing Data Using a Particle Filter}.
\newblock In \emph{Advanced Concepts for Intelligent Vision Systems}, 2006.

\bibitem[Fu et~al.(2011)Fu, Chen, Chou, Tsai, and Lee]{fu+2011vcip}
Tsung-Sheng Fu, Hua-Tsung Chen, Chien-Li Chou, Wen-Jiin Tsai, and Suh-Yin Lee.
\newblock {Screen-strategy Analysis in Broadcast Basketball Video using Player Tracking}.
\newblock In \emph{2011 Visual Communications and Image Processing (VCIP)}, 2011.

\bibitem[Ghasemzadeh et~al.(2021)Ghasemzadeh, Zandycke, Istasse, Sayez, Moshtaghpour, and Vleeschouwer]{ghasemzadeh+2021bmvc}
Seyed~Abolfazl Ghasemzadeh, Gabriel Zandycke, Maxime Istasse, Niels Sayez, Amirafshar Moshtaghpour, and Christophe Vleeschouwer.
\newblock {DeepSportLab: a Unified Framework for Ball Detection, Player Instance Segmentation and Pose Estimation in Team Sports Scenes}.
\newblock In \emph{BMVC}, 2021.

\bibitem[Glover and Kaelbling(2014)]{glover+2014icra}
Jared Glover and Leslie~Pack Kaelbling.
\newblock {Tracking the Spin on a Ping Pong Ball with the Quaternion Bingham Filter}.
\newblock In \emph{2014 IEEE International Conference on Robotics and Automation (ICRA)}, 2014.

\bibitem[He et~al.(2016)He, Zhang, Ren, and Sun]{he+2016cvpr}
Kaiming He, Xiangyu Zhang, Shaoqing Ren, and Jian Sun.
\newblock {Deep Residual Learning for Image Recognition}.
\newblock In \emph{2016 IEEE Conference on Computer Vision and Pattern Recognition (CVPR)}, 2016.

\bibitem[Huang et~al.(2011)Huang, Cox, Yan, de~Campos, Windridge, Kittler, and Christmas]{huang+2011avsp}
Qiang Huang, Stephen~J. Cox, Fei Yan, Te{\'o}filo~Em{\'i}dio de~Campos, David Windridge, Josef Kittler, and William~J. Christmas.
\newblock {Improved Detection of Ball Hit Events in a Tennis Game using Multimodal Information}.
\newblock In \emph{AVSP}, 2011.

\bibitem[Huang et~al.(2012{\natexlab{a}})Huang, Cox, Zhou, and Xie]{huang+2012apsipa}
Qiang Huang, Stephen Cox, Xiangzeng Zhou, and Lei Xie.
\newblock {Detection of Ball Hits in a Tennis Game using Audio and Visual Information}.
\newblock In \emph{Proceedings of The 2012 Asia Pacific Signal and Information Processing Association Annual Summit and Conference}, 2012{\natexlab{a}}.

\bibitem[Huang et~al.(2012{\natexlab{b}})Huang, Chen, Chiu, Yi, Lin, Yeh, and Kuo]{huang+2012icppw}
Yi-Chen Huang, Tsung-Long Chen, Bo-Chun Chiu, Chih-Wei Yi, Chung-Wei Lin, Yu-Jung Yeh, and Lun-Chia Kuo.
\newblock {Calculate Golf Swing Trajectories from IMU Sensing Data}.
\newblock In \emph{2012 41st International Conference on Parallel Processing Workshops}, 2012{\natexlab{b}}.

\bibitem[Huang et~al.(2008)Huang, Llach, and Zhang]{huang+2008icpr}
Yu~Huang, Joan Llach, and Chao Zhang.
\newblock {A Method of Small Object Detection and Tracking Based on Particle Filters}.
\newblock In \emph{2008 19th International Conference on Pattern Recognition}, 2008.

\bibitem[Huang et~al.(2019)Huang, Liao, Chen, İk, and Peng]{huang+2019avss}
Yu-Chuan Huang, I-No Liao, Ching-Hsuan Chen, Tsì-Uí İk, and Wen-Chih Peng.
\newblock {TrackNet: A Deep Learning Network for Tracking High-speed and Tiny Objects in Sports Applications}.
\newblock In \emph{2019 16th IEEE International Conference on Advanced Video and Signal Based Surveillance (AVSS)}, 2019.

\bibitem[Ibrahim et~al.(2016)Ibrahim, Muralidharan, Deng, Vahdat, and Mori]{Ibrahim+cvpr2016}
Mostafa~S. Ibrahim, Srikanth Muralidharan, Zhiwei Deng, Arash Vahdat, and Greg Mori.
\newblock {A Hierarchical Deep Temporal Model for Group Activity Recognition}.
\newblock In \emph{2016 IEEE Conference on Computer Vision and Pattern Recognition (CVPR)}, 2016.

\bibitem[Ishii et~al.(2007)Ishii, Kitahara, Kameda, and Ohta]{ishii+2007pcm}
Norihiro Ishii, Itaru Kitahara, Yoshinari Kameda, and Yuichi Ohta.
\newblock {3D Tracking of a Soccer Ball Using Two Synchronized Cameras}.
\newblock In \emph{Advances in Multimedia Information Processing -- PCM 2007}, 2007.

\bibitem[Kamble et~al.(2019)Kamble, Keskar, and Bhurchandi]{kamble+2018oer}
Paresh~R. Kamble, Avinash~G. Keskar, and Kishor~M. Bhurchandi.
\newblock {A Deep Learning Ball Tracking System in Soccer Videos}.
\newblock \emph{Opto-Electronics Review}, 2019.

\bibitem[Kim and Kim(2009)]{kim+2009cgiv}
Jong-Yun Kim and Tae-Yong Kim.
\newblock {Soccer Ball Tracking Using Dynamic Kalman Filter with Velocity Control}.
\newblock In \emph{2009 Sixth International Conference on Computer Graphics, Imaging and Visualization}, 2009.

\bibitem[Kingma and Ba(2015)]{kingma+2015iclr}
Diederik~P. Kingma and Jimmy Ba.
\newblock {Adam: {A} Method for Stochastic Optimization}.
\newblock In \emph{3rd International Conference on Learning Representations, {ICLR} 2015, San Diego, CA, USA, May 7-9, 2015, Conference Track Proceedings}, 2015.

\bibitem[Kittler et~al.(2005)Kittler, Christmas, Kostin, Yan, Kolonias, and Windridge]{kittler+2005ia}
Josef. Kittler, William~J. Christmas, Alexey Kostin, Fei. Yan, Ilias Kolonias, and David Windridge.
\newblock {A Memory Architecture and Contextual Reasoning Framework for Cognitive Vision}.
\newblock In \emph{Image Analysis}, 2005.

\bibitem[Kolonias et~al.(2007)Kolonias, Kittler, Christmas, and Yan]{kittler+2007iciap}
Ilias Kolonias, J.~Kittler, William Christmas, and Fly Yan.
\newblock Improving the accuracy of automatic tennis video annotation by high level grammar.
\newblock 2007.

\bibitem[Komorowski et~al.(2019)Komorowski, Kurzejamski, and Sarwas]{komorowski+2019mva}
Jacek Komorowski, Grzegorz Kurzejamski, and Grzegorz Sarwas.
\newblock {BallTrack: Football Ball Tracking for Real-time CCTV Systems}.
\newblock In \emph{2019 16th International Conference on Machine Vision Applications (MVA)}, 2019.

\bibitem[Komorowski. et~al.(2019)Komorowski., Kurzejamski., and Sarwas.]{komorowski+2019visapp}
Jacek Komorowski., Grzegorz Kurzejamski., and Grzegorz Sarwas.
\newblock {DeepBall: Deep Neural-Network Ball Detector}.
\newblock In \emph{Proceedings of the 14th International Joint Conference on Computer Vision, Imaging and Computer Graphics Theory and Applications - Volume 5: VISAPP,}, 2019.

\bibitem[Komorowski et~al.(2020)Komorowski, Kurzejamski, and Sarwas]{komorowski+2020visapp}
Jacek Komorowski, Grzegorz Kurzejamski, and Grzegorz Sarwas.
\newblock {Foot{A}nd{B}all: Integrated Player and Ball Detector}.
\newblock In \emph{Proceedings of the 15th International Joint Conference on Computer Vision, Imaging and Computer Graphics Theory and Applications - Volume 5: VISAPP,}, 2020.

\bibitem[Lepetit et~al.(2003)Lepetit, Shahrokni, and Fua]{lepetit+2003cvpr}
Vincent Lepetit, Ali Shahrokni, and Pascal Fua.
\newblock {Robust Data Association for Online Application}.
\newblock In \emph{2003 IEEE Computer Society Conference on Computer Vision and Pattern Recognition, 2003. Proceedings.}, 2003.

\bibitem[Li et~al.(2020{\natexlab{a}})Li, Wang, Hu, Li, Tang, and Yang]{li+2020arxiv}
Xiang Li, Wenhai Wang, Xiaolin Hu, Jun Li, Jinhui Tang, and Jian Yang.
\newblock {Generalized Focal Loss V2: Learning Reliable Localization Quality Estimation for Dense Object Detection}.
\newblock \emph{arXiv preprint}, 2020{\natexlab{a}}.

\bibitem[Li et~al.(2020{\natexlab{b}})Li, Wang, Wu, Chen, Hu, Li, Tang, and Yang]{li+2020neurips}
Xiang Li, Wenhai Wang, Lijun Wu, Shuo Chen, Xiaolin Hu, Jun Li, Jinhui Tang, and Jian Yang.
\newblock {Generalized Focal Loss: Learning Qualified and Distributed Bounding Boxes for Dense Object Detection}.
\newblock In \emph{NeurIPS}, 2020{\natexlab{b}}.

\bibitem[Li et~al.(2005)Li, Dore, and Orwell]{li+2005avss}
Yan Li, Alessio Dore, and James Orwell.
\newblock {Evaluating the Performance of Systems for Tracking Football Players and Ball}.
\newblock In \emph{IEEE Conference on Advanced Video and Signal Based Surveillance, 2005.}, 2005.

\bibitem[Liang et~al.(2005)Liang, Liu, Huang, and Gao]{liang+2005pcm}
Dawei Liang, Yang Liu, Qingming Huang, and Wen Gao.
\newblock {A Scheme for Ball Detection and Tracking in Broadcast Soccer Video}.
\newblock In \emph{Advances in Multimedia Information Processing - PCM 2005}, 2005.

\bibitem[Liang et~al.(2007)Liang, Huang, Liu, Zhu, and Gao]{liang+2007tce}
Dawei Liang, Qingming Huang, Yang Liu, Guangyu Zhu, and Wen Gao.
\newblock {Video2Cartoon: A System for Converting Broadcast Soccer Video into 3D Cartoon Animation}.
\newblock \emph{IEEE Transactions on Consumer Electronics}, 2007.

\bibitem[Lin et~al.(2017)Lin, Goyal, Girshick, He, and Dollár]{lin+2017iccv}
Tsung-Yi Lin, Priya Goyal, Ross Girshick, Kaiming He, and Piotr Dollár.
\newblock {Focal Loss for Dense Object Detection}.
\newblock In \emph{2017 IEEE International Conference on Computer Vision (ICCV)}, 2017.

\bibitem[Liu and Wang(2022)]{liu+2022cvprw}
Paul Liu and Jui-Hsien Wang.
\newblock {MonoTrack: Shuttle Trajectory Reconstruction From Monocular Badminton Video}.
\newblock In \emph{Proceedings of the IEEE/CVF Conference on Computer Vision and Pattern Recognition (CVPR) Workshops}, 2022.

\bibitem[Liu et~al.(2016)Liu, Anguelov, Erhan, Szegedy, Reed, Fu, and Berg]{liu+2016eccv}
Wei Liu, Dragomir Anguelov, Dumitru Erhan, Christian Szegedy, Scott Reed, Cheng-Yang Fu, and Alexander~C. Berg.
\newblock {SSD: Single Shot MultiBox Detector}.
\newblock In \emph{Computer Vision -- ECCV 2016}, 2016.

\bibitem[Liu et~al.(2006)Liu, Liang, Huang, and Gao]{liu+2006ivc}
Yang Liu, Dawei Liang, Qingming Huang, and Wen Gao.
\newblock {Extracting 3D Information from Broadcast Soccer Video}.
\newblock \emph{Image and Vision Computing}, 2006.

\bibitem[Lyu et~al.(2015)Lyu, Liu, Li, and Chen]{lyu+2015icia}
Congyi Lyu, Yunhui Liu, Bing Li, and Haoyao Chen.
\newblock {Multi-feature based High-speed Ball Shape Target Tracking}.
\newblock In \emph{2015 IEEE International Conference on Information and Automation}, 2015.

\bibitem[Lyu et~al.(2017)Lyu, Liu, Jiang, Li, and Chen]{lyu+2017ijsr}
Congyi Lyu, Yunhui Liu, Xin Jiang, Peng Li, and Haoyao Chen.
\newblock {High-Speed Object Tracking with Its Application in Golf Playing}.
\newblock \emph{International Journal of Social Robotics}, 2017.

\bibitem[M. and Pati(2015)]{rao+2015iccsp}
Upendra~Rao M. and Umesh~C. Pati.
\newblock {A Novel Algorithm for Detection of Soccer Ball and Player}.
\newblock In \emph{2015 International Conference on Communications and Signal Processing (ICCSP)}, 2015.

\bibitem[Misu et~al.(2007)Misu, Matsui, Naemura, Fujii, and Yagi]{misu+2007icassp}
Toshihiko Misu, Atsushi Matsui, Masahide Naemura, Mahito Fujii, and Nobuyuki Yagi.
\newblock {Distributed Particle Filtering for Multiocular Soccer-Ball Tracking}.
\newblock In \emph{2007 IEEE International Conference on Acoustics, Speech and Signal Processing - ICASSP '07}, 2007.

\bibitem[Miura et~al.(2009)Miura, Shimawaki, Sakiyama, and Shirai]{miura+2009cviu}
Jun Miura, Takumi Shimawaki, Takuro Sakiyama, and Yoshiaki Shirai.
\newblock {Ball Route Estimation under Heavy Occlusion in Broadcast Soccer Video}.
\newblock \emph{Computer Vision and Image Understanding}, 2009.

\bibitem[Myint et~al.(2015)Myint, Wong, Dooley, and Hopgood]{myint+2015mva}
Hnin Myint, Patrick Wong, Laurence Dooley, and Adrian Hopgood.
\newblock {Tracking a Table Tennis Ball for Umpiring Purposes}.
\newblock In \emph{2015 14th IAPR International Conference on Machine Vision Applications (MVA)}, 2015.

\bibitem[O~Conaire et~al.(2009)O~Conaire, Kelly, Connaghan, and O'Connor]{conaire+2009icdsp}
Ciaran O~Conaire, Philip Kelly, Damien Connaghan, and Noel~E. O'Connor.
\newblock {TennisSense: A Platform for Extracting Semantic Information from Multi-Camera Tennis Data}.
\newblock In \emph{2009 16th International Conference on Digital Signal Processing}, 2009.

\bibitem[Ohno et~al.(1999)Ohno, Miura, and Shirai]{ohno+1999mfi}
Yoshinori. Ohno, Jun. Miura, and Yoshiaki Shirai.
\newblock {Tracking Players and a Ball in Soccer Games}.
\newblock In \emph{Proceedings. 1999 IEEE/SICE/RSJ. International Conference on Multisensor Fusion and Integration for Intelligent Systems. MFI'99 (Cat. No.99TH8480)}, 1999.

\bibitem[Ohno et~al.(2000)Ohno, Miura, and Shirai]{ohno+2000icpr}
Yoshinori Ohno, Jun Miura, and Yoshiaki Shirai.
\newblock {Tracking Players and Estimation of the 3D Position of a Ball in Soccer Games}.
\newblock In \emph{Proceedings 15th International Conference on Pattern Recognition. ICPR-2000}, 2000.

\bibitem[Pallavi et~al.(2008)Pallavi, Mukherjee, Majumdar, and Sural]{pallavi+2008jvcir}
V.~Pallavi, Jayanta Mukherjee, Arun~K. Majumdar, and Shamik Sural.
\newblock {Ball Detection from Broadcast Soccer Videos using Static and Dynamic Features}.
\newblock \emph{Journal of Visual Communication and Image Representation}, 2008.

\bibitem[Perez et~al.(2022)Perez, Liu, and Kot]{perez+2022pr}
Mauricio Perez, Jun Liu, and Alex~C. Kot.
\newblock {Skeleton-based Relational Reasoning for Group Activity Analysis}.
\newblock \emph{Pattern Recognition}, 2022.

\bibitem[Pingali et~al.(2000)Pingali, Opalach, and Jean]{pingali+2000icpr}
Gopal Pingali, Agata Opalach, and Yves~D. Jean.
\newblock {Ball Tracking and Virtual Replays for Innovative Tennis Broadcasts}.
\newblock In \emph{Proceedings 15th International Conference on Pattern Recognition. ICPR-2000}, 2000.

\bibitem[Rea et~al.(2004)Rea, Dahyot, and Kokaram]{rea+2004ivr}
Niall Rea, Rozenn Dahyot, and Anil Kokaram.
\newblock {Semantic Event Detection in Sports Through Motion Understanding}.
\newblock In \emph{Image and Video Retrieval}, 2004.

\bibitem[Ren et~al.(2006)Ren, Orwell, and Jones]{ren+2006eccvw}
Jinchang Ren, James Orwell, and Graeme~A. Jones.
\newblock {Generating Ball Trajectory in Soccer Video Sequences}.
\newblock In \emph{ECCV Workshops}, 2006.

\bibitem[Ren et~al.(2008)Ren, Orwell, Jones, and Xu]{ren+2008tcsvt}
Jinchang Ren, James Orwell, Graeme~A. Jones, and Ming Xu.
\newblock {Real-Time Modeling of 3-D Soccer Ball Trajectories From Multiple Fixed Cameras}.
\newblock \emph{IEEE Transactions on Circuits and Systems for Video Technology}, 2008.

\bibitem[Ren et~al.(2009)Ren, Orwell, Jones, and Xu]{ren+2009cviu}
Jinchang Ren, James Orwell, Graeme~A. Jones, and Ming Xu.
\newblock {Tracking the Soccer Ball using Multiple Fixed Cameras}.
\newblock \emph{Computer Vision and Image Understanding}, 2009.

\bibitem[Renò et~al.(2016)Renò, Mosca, Nitti, Guaragnella, D'Orazio, and Stella]{reno+2016tishw}
Vito Renò, Nicola Mosca, Massimiliano Nitti, Cataldo Guaragnella, Tiziana D'Orazio, and Ettore Stella.
\newblock {Real-time Tracking of a Tennis Ball by Combining 3D Data and Domain Knowledge}.
\newblock In \emph{2016 1st International Conference on Technology and Innovation in Sports, Health and Wellbeing (TISHW)}, 2016.

\bibitem[Renò et~al.(2018)Renò, Mosca, Marani, Nitti, D'Orazio, and Stella]{reno+2018cvprw}
Vito Renò, Nicola Mosca, Roberto Marani, Massimiliano Nitti, Tiziana D'Orazio, and Ettore Stella.
\newblock {Convolutional Neural Networks Based Ball Detection in Tennis Games}.
\newblock In \emph{2018 IEEE/CVF Conference on Computer Vision and Pattern Recognition Workshops (CVPRW)}, 2018.

\bibitem[Ronneberger et~al.(2015)Ronneberger, Fischer, and Brox]{ronneberger+2015miccai}
Olaf Ronneberger, Philipp Fischer, and Thomas Brox.
\newblock {U-Net: Convolutional Networks for Biomedical Image Segmentation}.
\newblock In \emph{Medical Image Computing and Computer-Assisted Intervention -- MICCAI 2015}, 2015.

\bibitem[Sarkar et~al.(2019)Sarkar, Chakrabarti, and Prasad~Mukherjee]{sarkar+2019cvprw}
Saikat Sarkar, Amlan Chakrabarti, and Dipti Prasad~Mukherjee.
\newblock {Generation of Ball Possession Statistics in Soccer Using Minimum-Cost Flow Network}.
\newblock In \emph{Proceedings of the IEEE/CVF Conference on Computer Vision and Pattern Recognition (CVPR) Workshops}, 2019.

\bibitem[Shimawaki et~al.(2006)Shimawaki, Sakiyama, Miura, and Shirai]{shimawaki+2006icpr}
Takumi Shimawaki, Takuro Sakiyama, Jun Miura, and Yoshiaki Shirai.
\newblock {Estimation of Ball Route under Overlapping with Players and Lines in Soccer Video Image Sequence}.
\newblock In \emph{18th International Conference on Pattern Recognition (ICPR'06)}, 2006.

\bibitem[Shum and Komura(2004)]{shum+2004icme}
Hubert P.~H. Shum and Taku Komura.
\newblock {A Spatiotemporal Approach to Extract the 3D Trajectory of the Baseball from a Single View Video Sequence}.
\newblock In \emph{2004 IEEE International Conference on Multimedia and Expo (ICME) (IEEE Cat. No.04TH8763)}, 2004.

\bibitem[Sun et~al.(2020)Sun, Lin, Chuang, Hsu, Yu, Chung, and İk]{sun+2020icpai}
Nien-En Sun, Yu-Ching Lin, Shao-Ping Chuang, Tzu-Han Hsu, Dung-Ru Yu, Ho-Yi Chung, and Tsì-Uí İk.
\newblock {TrackNetV2: Efficient Shuttlecock Tracking Network}.
\newblock In \emph{2020 International Conference on Pervasive Artificial Intelligence (ICPAI)}, 2020.

\bibitem[Taghanaki et~al.(2019)Taghanaki, Zheng, {Kevin Zhou}, Georgescu, Sharma, Xu, Comaniciu, and Hamarneh]{taghanaki+cmig}
Saeid~Asgari Taghanaki, Yefeng Zheng, S.~{Kevin Zhou}, Bogdan Georgescu, Puneet Sharma, Daguang Xu, Dorin Comaniciu, and Ghassan Hamarneh.
\newblock {Combo Loss: Handling Input and Output Imbalance in Multi-organ Segmentation}.
\newblock \emph{Computerized Medical Imaging and Graphics}, 2019.

\bibitem[Teachabarikiti et~al.(2010)Teachabarikiti, Chalidabhongse, and Thammano]{teachabarikiti+2010icca}
Kosit Teachabarikiti, Thanarat~H. Chalidabhongse, and Arit Thammano.
\newblock {Players Tracking and Ball Detection for an Automatic Tennis Video Annotation}.
\newblock In \emph{2010 11th International Conference on Control Automation Robotics \& Vision}, 2010.

\bibitem[Theagarajan et~al.(2018)Theagarajan, Pala, Zhang, and Bhanu]{theagarajan+cvpr2018}
Rajkumar Theagarajan, Federico Pala, Xiu Zhang, and Bir Bhanu.
\newblock {Soccer: Who Has the Ball? Generating Visual Analytics and Player Statistics}.
\newblock In \emph{2018 IEEE/CVF Conference on Computer Vision and Pattern Recognition Workshops (CVPRW)}, 2018.

\bibitem[Tong et~al.(2004)Tong, Lu, and Liu]{tong+2004icpr}
Xiao-Feng Tong, Han-Qing Lu, and Qing-Shan Liu.
\newblock {An Effective and Fast Soccer Ball Detection and Tracking Method}.
\newblock In \emph{Proceedings of the 17th International Conference on Pattern Recognition, 2004. ICPR 2004.}, 2004.

\bibitem[Van~Zandycke and De~Vleeschouwer(2019)]{zandycke+2019mmsports}
Gabriel Van~Zandycke and Christophe De~Vleeschouwer.
\newblock {Real-Time CNN-Based Segmentation Architecture for Ball Detection in a Single View Setup}.
\newblock In \emph{Proceedings Proceedings of the 2nd International Workshop on Multimedia Content Analysis in Sports}, 2019.

\bibitem[Voeikov et~al.(2020)Voeikov, Falaleev, and Baikulov]{voeikov+2020cvprw}
Roman Voeikov, Nikolay Falaleev, and Ruslan Baikulov.
\newblock {TTNet: Real-Time Temporal and Spatial Video Analysis of Table Tennis}.
\newblock In \emph{Proceedings of the IEEE/CVF Conference on Computer Vision and Pattern Recognition (CVPR) Workshops}, June 2020.

\bibitem[Wang et~al.(2021)Wang, Sun, Cheng, Jiang, Deng, Zhao, Liu, Mu, Tan, Wang, Liu, and Xiao]{wang+2021tpami}
Jingdong Wang, Ke~Sun, Tianheng Cheng, Borui Jiang, Chaorui Deng, Yang Zhao, Dong Liu, Yadong Mu, Mingkui Tan, Xinggang Wang, Wenyu Liu, and Bin Xiao.
\newblock {Deep High-Resolution Representation Learning for Visual Recognition}.
\newblock \emph{IEEE Transactions on Pattern Analysis and Machine Intelligence}, 2021.

\bibitem[Wang et~al.(2022)Wang, Shuai, Chang, and Peng]{wang+2022aaai}
Wei-Yao Wang, Hong-Han Shuai, Kai-Shiang Chang, and Wen-Chih Peng.
\newblock {ShuttleNet: Position-Aware Fusion of Rally Progress and Player Styles for Stroke Forecasting in Badminton}.
\newblock \emph{Proceedings of the AAAI Conference on Artificial Intelligence}, 2022.

\bibitem[Wang et~al.(2014)Wang, Ablavsky, Shitrit, and Fua]{wang+2014cviu}
Xinchao Wang, Vitaly Ablavsky, Horesh~Ben Shitrit, and Pascal Fua.
\newblock {Take Your Eyes Off the Ball: Improving Ball-tracking by Focusing on Team Play}.
\newblock \emph{Computer Vision and Image Understanding}, 2014.

\bibitem[Wang et~al.(2016)Wang, Cheng, Ikoma, Honda, and Ikenaga]{yuan+2016scis}
Yuan Wang, Xina Cheng, Norikazu Ikoma, Masaaki Honda, and Takeshi Ikenaga.
\newblock {Motion Prejudgment Dependent Mixture System Noise in System Model for Tennis Ball 3D Position Tracking by Particle Filter}.
\newblock In \emph{2016 Joint 8th International Conference on Soft Computing and Intelligent Systems (SCIS) and 17th International Symposium on Advanced Intelligent Systems (ISIS)}, 2016.

\bibitem[Wong and Dooley(2010)]{wong+2010ics}
Kam Cheung~Patrik Wong and Laurence~S. Dooley.
\newblock {High-motion Table Tennis Ball Tracking for Umpiring Applications}.
\newblock In \emph{IEEE 10th INTERNATIONAL CONFERENCE ON SIGNAL PROCESSING PROCEEDINGS}, 2010.

\bibitem[Wu et~al.(2020)Wu, Xu, Liang, Mei, and Peng]{wu+2020iet}
Wanneng Wu, Min Xu, Qiaokang Liang, Li~Mei, and Yu~Peng.
\newblock {Multi-camera 3D Ball Tracking Framework for Sports Video}.
\newblock \emph{IET Image Processing}, 2020.

\bibitem[Yan et~al.(2005)Yan, Christmas, and Kittler]{yan+2005bmvc}
Fei Yan, William~J. Christmas, and Josef Kittler.
\newblock {A Tennis Ball Tracking Algorithm for Automatic Annotation of Tennis Match}.
\newblock In \emph{BMVC}, 2005.

\bibitem[Yan et~al.(2008)Yan, Christmas, and Kittler]{yan+2008tpami}
Fei Yan, William Christmas, and Josef Kittler.
\newblock {Layered Data Association Using Graph-Theoretic Formulation with Application to Tennis Ball Tracking in Monocular Sequences}.
\newblock \emph{IEEE Transactions on Pattern Analysis and Machine Intelligence}, 2008.

\bibitem[Yan et~al.(2014)Yan, Christmas, and Kittler]{yan+2014cvs}
Fei Yan, William Christmas, and Josef Kittler.
\newblock \emph{{Ball Tracking for Tennis Video Annotation}}.
\newblock 2014.

\bibitem[Yan et~al.(2020)Yan, Xie, Tang, Shu, and Tian]{yan+2020eccv}
Rui Yan, Lingxi Xie, Jinhui Tang, Xiangbo Shu, and Qi~Tian.
\newblock {Social Adaptive Module for Weakly-Supervised Group Activity Recognition}.
\newblock In \emph{Computer Vision -- ECCV 2020}, 2020.

\bibitem[Yu et~al.(2021)Yu, Xiao, Gao, Yuan, Zhang, Sang, and Wang]{yu+2021cvpr}
Changqian Yu, Bin Xiao, Changxin Gao, Lu~Yuan, Lei Zhang, Nong Sang, and Jingdong Wang.
\newblock {Lite-HRNet: A Lightweight High-Resolution Network}.
\newblock In \emph{2021 IEEE/CVF Conference on Computer Vision and Pattern Recognition (CVPR)}, 2021.

\bibitem[Yu et~al.(2007{\natexlab{a}})Yu, Tang, Wang, and Shi]{yu+2007avc}
Junqing Yu, Yang Tang, Zhifang Wang, and Lejiang Shi.
\newblock {Playfield and Ball Detection in Soccer Video}.
\newblock In \emph{Advances in Visual Computing}, 2007{\natexlab{a}}.

\bibitem[Yu et~al.(2003{\natexlab{a}})Yu, Tian, and Wan]{yu+2003icme}
Xinguo Yu, Qi~Tian, and Kong~Wah Wan.
\newblock {A Novel Ball Detection Framework for Real Soccer Video}.
\newblock In \emph{2003 International Conference on Multimedia and Expo. ICME '03. Proceedings (Cat. No.03TH8698)}, 2003{\natexlab{a}}.

\bibitem[Yu et~al.(2003{\natexlab{b}})Yu, Xu, Tian, and Leong]{yu+2003icme2}
Xinguo Yu, Changshen Xu, Qi~Tian, and Hon~Wai Leong.
\newblock {A Ball Tracking Framework for Broadcast Soccer Video}.
\newblock In \emph{2003 International Conference on Multimedia and Expo. ICME '03. Proceedings (Cat. No.03TH8698)}, 2003{\natexlab{b}}.

\bibitem[Yu et~al.(2003{\natexlab{c}})Yu, Xu, Leong, Tian, Tang, and Wan]{yu+2003acmmm}
Xinguo Yu, Changsheng Xu, Hon~Wai Leong, Qi~Tian, Qing Tang, and Kong~Wah Wan.
\newblock {Trajectory-Based Ball Detection and Tracking with Applications to Semantic Analysis of Broadcast Soccer Video}.
\newblock In \emph{Proceedings of the Eleventh ACM International Conference on Multimedia}, 2003{\natexlab{c}}.

\bibitem[Yu et~al.(2004{\natexlab{a}})Yu, Sim, Wang, and Cheong]{yu+2004icip}
Xinguo Yu, Chern-Horng Sim, Jenny~R. Wang, and Loong~Fah Cheong.
\newblock {A Trajectory-based Ball Detection and Tracking Algorithm in Broadcast Tennis Video}.
\newblock In \emph{2004 International Conference on Image Processing, 2004. ICIP '04.}, 2004{\natexlab{a}}.

\bibitem[Yu et~al.(2004{\natexlab{b}})Yu, Yan, Hay, and Leong]{yu+2004acmmm}
Xinguo Yu, Xin Yan, Tze~Sen Hay, and Hon~Wai Leong.
\newblock {3D Reconstruction and Enrichment of Broadcast Soccer Video}.
\newblock In \emph{Proceedings of the 12th Annual ACM International Conference on Multimedia}, 2004{\natexlab{b}}.

\bibitem[Yu et~al.(2005)Yu, Hay, Yan, and Chng]{yu+2005icme}
Xinguo Yu, Tze~Sen Hay, Xin Yan, and E.~Chng.
\newblock {A Player-Possession Acquisition System for Broadcast Soccer Video}.
\newblock In \emph{2005 IEEE International Conference on Multimedia and Expo}, 2005.

\bibitem[Yu et~al.(2006)Yu, Leong, Xu, and Tian]{yu+2006tmm}
Xinguo Yu, Hon~Wai Leong, Changsheng Xu, and Qi~Tian.
\newblock {Trajectory-Based Ball Detection and Tracking in Broadcast Soccer Video}.
\newblock \emph{IEEE Transactions on Multimedia}, 2006.

\bibitem[Yu et~al.(2007{\natexlab{b}})Yu, Jiang, and Ang]{yu+2007vcip}
Xinguo Yu, Nianjuan Jiang, and Ee~Luang Ang.
\newblock {Trajectory-based Ball Detection and Tracking with Aid of Homography in Broadcast Tennis Video}.
\newblock In \emph{Visual Communications and Image Processing 2007}, 2007{\natexlab{b}}.

\bibitem[Yu et~al.(2007{\natexlab{c}})Yu, Tu, and Ang]{yu+2007icme}
Xinguo Yu, Xiaoying Tu, and Ee~Luang Ang.
\newblock {Trajectory-Based Ball Detection and Tracking in Broadcast Soccer Video with the Aid of Camera Motion Recovery}.
\newblock In \emph{2007 IEEE International Conference on Multimedia and Expo}, 2007{\natexlab{c}}.

\bibitem[Yu et~al.(2009)Yu, Jiang, Cheong, Leong, and Yan]{yu+2009cviu}
Xinguo Yu, Nianjuan Jiang, Loong-Fah Cheong, Hon~Wai Leong, and Xin Yan.
\newblock {Automatic Camera Calibration of Broadcast Tennis Video with Applications to 3D Virtual Content Insertion and Ball Detection and Tracking}.
\newblock \emph{Computer Vision and Image Understanding}, 2009.

\bibitem[Zaveri et~al.(2004)Zaveri, Merchant, and Desai]{zaveri+2004icme}
Mukesh~A. Zaveri, Shabbir~N. Merchant, and Uday~B. Desai.
\newblock {Small and Fast Moving Object Detection and Tracking in Sports Video Sequences}.
\newblock In \emph{2004 IEEE International Conference on Multimedia and Expo (ICME) (IEEE Cat. No.04TH8763)}, 2004.

\bibitem[Zeiler(2012)]{zeiler2012arxiv}
Matthew~D. Zeiler.
\newblock {ADADELTA: An Adaptive Learning Rate Method}, 2012.

\bibitem[Zhang et~al.(2011)Zhang, Wei, Yu, and Zhong]{zhang+2011zus}
Yuan-hui Zhang, Wei Wei, Dan Yu, and Cong-wei Zhong.
\newblock {A Tracking and Predicting Scheme for Ping Pong Robot}.
\newblock \emph{Journal of Zhejiang University SCIENCE C}, 2011.

\bibitem[Zhang et~al.(2010)Zhang, Xu, and Tan]{zhang+2010tim}
Zhengtao Zhang, De~Xu, and Min Tan.
\newblock {Visual Measurement and Prediction of Ball Trajectory for Table Tennis Robot}.
\newblock \emph{IEEE Transactions on Instrumentation and Measurement}, 2010.

\bibitem[Zhang et~al.(2022)Zhang, Wu, Qiu, Liang, and Li]{zhang+2022arxiv}
Zhewen Zhang, Fuliang Wu, Yuming Qiu, Jingdong Liang, and Shuiwang Li.
\newblock Tracking small and fast moving objects: A benchmark.
\newblock In \emph{Proceedings of the Asian Conference on Computer Vision (ACCV)}, 2022.

\bibitem[Zhao et~al.(2018)Zhao, Qi, Shen, Shi, and Jia]{zhao+2018eccv}
Hengshuang Zhao, Xiaojuan Qi, Xiaoyong Shen, Jianping Shi, and Jiaya Jia.
\newblock {ICNet for Real-Time Semantic Segmentation on High-Resolution Images}.
\newblock In \emph{Computer Vision -- ECCV 2018}, 2018.

\bibitem[Zhou et~al.(2013)Zhou, Huang, Xie, and Cox]{zhou+2013icassp}
Xiangzeng Zhou, Qiang Huang, Lei Xie, and Stephen Cox.
\newblock {A Two Layered Data Association Approach for Ball Tracking}.
\newblock In \emph{2013 IEEE International Conference on Acoustics, Speech and Signal Processing}, 2013.

\bibitem[Zhou et~al.(2015)Zhou, Xie, Huang, Cox, and Zhang]{zhou+2015tmm}
Xiangzeng Zhou, Lei Xie, Qiang Huang, Stephen~J. Cox, and Yanning Zhang.
\newblock {Tennis Ball Tracking Using a Two-Layered Data Association Approach}.
\newblock \emph{IEEE Transactions on Multimedia}, 2015.

\bibitem[Zhu et~al.(2008)Zhu, Xu, Zhang, Huang, and Lu]{zhu+2008civr}
Guangyu Zhu, Changsheng Xu, Yi~Zhang, Qingming Huang, and Hanqing Lu.
\newblock {Event Tactic Analysis Based on Player and Ball Trajectory in Broadcast Video}.
\newblock In \emph{Proceedings of the 2008 International Conference on Content-Based Image and Video Retrieval}, 2008.

\bibitem[Zhu et~al.(2009)Zhu, Xu, Huang, Rui, Jiang, Gao, and Yao]{zhu+2009tmm}
Guangyu Zhu, Changsheng Xu, Qingming Huang, Yong Rui, Shuqiang Jiang, Wen Gao, and Hongxun Yao.
\newblock {Event Tactic Analysis Based on Broadcast Sports Video}.
\newblock \emph{IEEE Transactions on Multimedia}, 2009.

\end{thebibliography}

\end{document}